\documentclass[10pt,twocolumn,letterpaper]{article}

\usepackage[cvpr]{cvpr}      %

\usepackage[dvipsnames]{xcolor}

\definecolor{cvprblue}{rgb}{0.21,0.49,0.74}
\usepackage[pagebackref,breaklinks,colorlinks,citecolor=cvprblue]{hyperref}
\usepackage{multirow}

\def\paperID{*****} %
\def\confName{3DV\xspace}
\def\confYear{2024\xspace}

\title{PlaNeRF: SVD Unsupervised 3D Plane Regularization for NeRF Large-Scale Urban Scene Reconstruction}

\author{Fusang Wang\\
Huawei Noah's Ark Lab\\ 
Shanghai Jiao Tong University\\
{\tt\small fusangwang@sjtu.edu.cn}
\and
Arnaud Louys\\
Huawei Noah's Ark Lab\\
{\tt\small arnaud.louys@gmail.com}
\and
Nathan Piasco\\
Huawei Noah's Ark Lab\\
{\tt\small nathan.piasco@huawei.com}
\and
Moussab Bennehar\\
Huawei Noah's Ark Lab\\
{\tt\small moussab.bennehar@huawei.com}
\and
Luis Rold\~ao\\
Huawei Noah's Ark Lab\\
{\tt\small luis.roldao@huawei.com}
\and
Dzmitry Tsishkou\\
Huawei Noah's Ark Lab\\
{\tt\small dzmitry.tsishkou@huawei.com}
}

\begin{document}
\twocolumn[{%
\renewcommand\twocolumn[1][]{#1}%
\maketitle

\begin{center}
	\centering
    \vspace{-0.7cm}
	\includegraphics[width=0.90\textwidth]{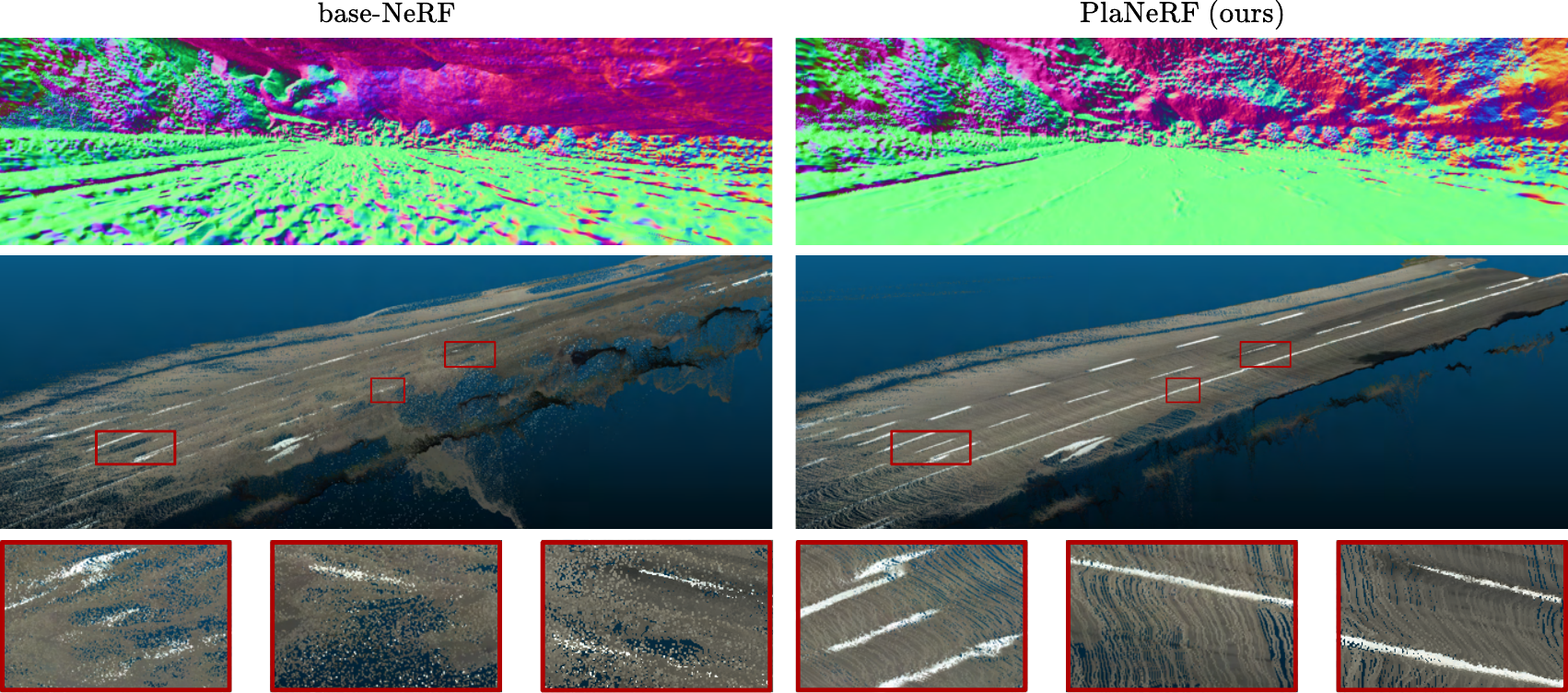}
	\captionof{figure}{While NeRF provides high quality renderings for views close to training poses, the rendering quality degrades drastically at extrapolated poses due to incorrect learned geometry. PlaNeRF results in more accurate geometry without 3D supervision, as shown by predicted normal maps (top) and point cloud (bottom).}
	\label{fig:teaser}
\end{center}
}]
\maketitle

\newcommand{\FW}[1]{\textcolor{orange}{[\textbf{FW}: #1]}}
\newcommand{\AL}[1]{\textcolor{purple}{[\textbf{AL}: #1]}}
\newcommand{\NP}[1]{\textcolor{cyan}{[\textbf{NP}: #1]}}
\newcommand{\MB}[1]{\textcolor{red}{[\textbf{MB}: #1]}}
\newcommand{\LR}[1]{\textcolor{blue}{[\textbf{LR}: #1]}}

\begin{abstract}
Neural Radiance Fields (NeRF) enable 3D scene reconstruction from 2D images and camera poses for Novel View Synthesis (NVS). Although NeRF can produce photorealistic results, it often suffers from overfitting to training views, leading to poor geometry reconstruction, especially in low-texture areas such as road surfaces in driving scenarios. This limitation restricts many important applications which require accurate geometry, such as extrapolated NVS, HD mapping, simulation and scene editing. 
To address this limitation, we propose a new method to improve NeRF's 3D structure using only RGB images and semantic maps. Our approach introduces a novel plane regularization based on Singular Value Decomposition (SVD), that does not rely on any geometric prior. In addition, we leverage the Structural Similarity Index Measure (SSIM) in patch-based loss design to properly initialize the volumetric representation of NeRF. Quantitative and qualitative results show that our method outperforms popular regularization approaches in accurate geometry reconstruction for large-scale outdoor scenes and achieves comparable rendering quality to SOTA methods on the KITTI-360 NVS benchmark.
\end{abstract}

\section{Introduction}

Neural Radiance Fields (NeRF) have recently emerged as an appealing solution for modeling 3D scenes from posed 2D images~\cite{mildenhall2021nerf}, finding increasing utility in many scenarios, notably in autonomous driving~\cite{herau2023moisst, moreau2022lens, rematas2022urban}. NeRF volumetric scene representation is obtained by optimizing an implicit model through differentiable ray marching using multi-view clues provided by a set of training images. Without explicit 3D supervision, neural field approaches suffer from the same drawbacks as standard multi-view stereo (MVS) methods for dense reconstruction~\cite{schonberger2016pixelwise}. For instance, they can easily overfit to training views, especially at low texture areas, resulting in poor geometry. This limitation hinders many important applications in autonomous driving and virtual reality such as HD mapping~\cite{wang2018lanenet, elhousni2020automatichd, paek2022rowhd}, view extrapolation~\cite{niemeyer2022regnerf, jain2021dietnerf}, scene editing~\cite{Ost2021neuralscene} or simulation~\cite{wu2023mars,yang2023unisim}, where accurate geometry for surface reconstruction is mandatory. Thus, learning good geometry becomes a challenging and crucial issue to improve the performance of NeRF for such applications.

Existing methods improve NeRF geometry by introducing more constraints into the optimization problem by using 3D supervision~\cite{rematas2022urban, azinovicneuralsdf, wang2022neuris, wang2021neus}. These solutions rely on LiDAR point clouds or well-reconstructed meshes, thus can fail on areas where 3D supervision is sparse and are sensitive to calibration errors. To overcome this limitation, recent trends aim to regularize NeRF geometry with less dependence on 3D information~\cite{guo2022manhattan, niemeyer2022regnerf, ehret2022diffnerf, structnerf, p2sdf}. Such methods impose priors on planar area and surface roughness and are able to learn accurate 3D structure. However, these priors do not hold in many scenarios and are limited to well-structured indoor scenes or small outdoor environments captured in controlled setups. Thus, improving the geometry reconstruction of large scenes without 3D guidance remains a major challenge.

In this paper, we present PlaNeRF, a new unsupervised regularization method \textit{without} explicit 3D supervision for learning correct geometry of outdoor scenes. Specifically, we introduce a new plane regularization method based on Singular Value Decomposition (SVD) that does not rely on any surface normal priors as it uses the initial geometry learnt by the NeRF. In order to initialize a proper geometry for our regularization term, we propose the use of the patch Structural Similarity Index Measure (SSIM) as supervision loss. We show that our method can lead to accurate geometry reconstruction in large unbounded urban scenes from autonomous driving datasets and achieves comparable rendering quality to SOTA Novel View Synthesis (NVS) methods in such scenarios. We summarize our contributions as follows:
    \begin{itemize}
        \item We propose a new SVD-based regularization method that improves NeRF geometry on large outdoor urban scenes without relying on explicit nor prior 3D information (Section \ref{sec:SVD_plane_regu}).
        \item We introduce a simple yet effective loss that minimizes the structural similarity index of image patches to properly initialize the NeRF volumetric density (Section \ref{sec:dssim_loss}).
        \item We perform an extensive analysis of existing geometry regularization methods applied on NeRF and demonstrate that our proposal outperforms existing solutions for accurate geometric outdoor scene reconstruction (Section \ref{sec:geometric_regu_comparison}).
        \item Results on the KITTI-360 NVS benchmark~\cite{liao2022kitti} show that our method achieves comparable performance to SoTA methods in terms of visual quality while improving the 3D geometric integrity.
    \end{itemize}

\section{Related Works}
In this section, we discuss previous works that target accurate 3D reconstruction from 2D images in outdoor scenes, with a special focus on surface and plane regularization techniques. We first introduce traditional SfM methods combined with MVS and their variants enhanced by deep learning and planar priors. We then present recent works on neural implicit representations of 3D scenes for surface reconstruction, focusing on different techniques relying on geometric priors. Finally, we discuss weakly-supervised or unsupervised geometry regularization techniques for 3D scene reconstruction.

\subsection{3D Reconstruction from 2D images}
Traditional structure-from-motion combined with multi-view stereo methods~\cite{schonberger2016colmap,schonberger2016pixelwise} have shown high accuracy but suffer from incomplete and sparse reconstruction results when few feature points are triangulated. To overcome these limitations, several methods combined with deep learning strategies have been proposed to improve different processes in the traditional MVS pipeline, including feature matching~\cite{leroy2018shape, ummenhofer2017demon}, dense depth estimation~\cite{zhang2022nerfusion, riegler2017octnetfusion} and depth fusion~\cite{yao2018mvsnet, yao2019recurrentmvsnet, yu2020fastmvsnet}. 
Another branch of work focuses on the construction of low texture planar surfaces and proposes to use a planar classifier~\cite{gallup2010planarMVS} to enforce a second order smoothness prior~\cite{woodford2009secondorder} or to introduce planar priors~\cite{xu2020planarpatchmatch, romanoni2019tapa, sun2021phi}.
\citet{xu2020planarpatchmatch} introduced a new multi-view matching cost function that combines planar constraints and photometric consistency that improves depth prediction in both planar and non-planar areas. In a concurrent work, \citet{sun2021phi} introduced predefined planar normals and propose a Markov Random Field based decision process that successfully reconstruct low texture planar surfaces. Having observed the effectiveness of incorporating a planar prior for low-texture reconstruction, we explore NeRF geometry regularization with a novel plane regularization technique based on singular value decomposition.

\subsection{NeRF for Surface Reconstruction}
Neural Radiance Fields, originally proposed by~\citet{mildenhall2021nerf}, formulate the 3D scene representation problem as an implicit neural function. Several works are proposed to optimize the rendering speed and the representability of NeRF. \citet{muller2022instant} use a multi-resolution hash table for the position encoding that greatly improves the rendering and convergence speed. In MipNeRF-360, \citet{barron2022mip360} redesign the radiance field structure as two MLPs, a smaller proposal MLP to produce guidance weights and a standard NeRF in charge of modeling the final weights and colors. Along with a better scene coordinate contraction and a novel ray density regularization loss, MipNeRF-360 is able to produce convincing results in outdoor scenes. Although the aforementioned solutions have shown great NVS performance in optimal conditions, they usually suffer from overfitting to training views and fail to generate extrapolated images in less favorable situations~\cite{zhang2022ray}.

To overcome this limitation, some approaches leverage 3D supervision signals such as LiDAR data, depth maps or normals from 3D meshes to constrain the original optimization problem~\cite{rematas2022urban, deng2022depth, wang2022neuris}. These approaches rely on well calibrated 3D ground truth and often fail when the supervision signal is weak. 
Conversely, \citet{wang2021neus} propose to rather optimize the signed distance function (SDF) outputted by the neural representation and succeeds in recovering more accurate geometry for indoor scenarios.
To further improve the quality of the recovered surface, geometric signal or strong surface normal hypotheses 
are employed to supervise the network training~\cite{yu2022monosdf, zou2022mononeuralfusion, guo2022manhattan}.
Given the well-defined surface property provided by the SDF, these methods recover high quality geometry but are usually limited to indoor scenes. 
In particular, MonoSDF~\cite{yu2022monosdf} uses monocular depth and normal cues from pre-trained Omnidata models to supervize NeRF geometry,
but its performance is highly dependent on the accuracy of the pre-trained model, which limits its ability to handle larger outdoor scenes. 
\citet{sun2022neural} introduce one of the few methods that successfully apply SDF scene modeling to outdoor scenarios. Their proposal benefit from a carefully designed multi-step sampling strategy and the use of appearance embeddings, resulting in slower training and requires a important view coverage. Meanwhile, the aforementioned method focuses on the reconstruction of outdoor landmarks, which is different from the complex urban scene commonly acquired by vehicle-mounted sensors.

\subsection{Plane Geometry Regularization with NeRF}
There is also a recent trend to regularize NeRF geometry with less dependency on 3D information by either introducing geometry regularization terms~\cite{ehret2022diffnerf, niemeyer2022regnerf}, leveraging planar priors~\cite{guo2022manhattan, structnerf, p2sdf} or employing frequency regularization~\cite{yang2023freenerf}.
FreeNeRF~\cite{yang2023freenerf} propose to gradually release high frequency features while training to improve NeRF geometry.
DiffNeRF~\cite{ehret2022diffnerf} introduces the use of differential geometry regularization for NeRF volumetric representation outperforming previous regularization methods for few-shot learning. However, it requires calculating higher order derivatives resulting in increased complexity and training times.
Both structNeRF~\cite{structnerf} and $P^2$SDF~\cite{p2sdf} use planar detectors to facilitate reconstructions on non-Manhattan indoor scenes. $P^2$SDF suggests a coarse to fine surface estimation strategy to regularize the SDF prediction on planar surfaces. StructNeRF proposes patch based multi-view consistency loss regularization enforcing points in planar region using planar constraints. 

In particular, RegNeRF~\cite{niemeyer2022regnerf} proposed a depth smoothness patch-based regularization method that pushes adjacent pixels to similar depth values. Such formulation leads to the prior of planes orthogonal to the center ray. To mitigate this behaviour, the authors propose to apply the depth smoothing regularization using an additional virtual camera generated throughout the 3D scenes. This solution produces smoother normal estimation at the cost of additional computational time, only if 
the positions of the virtual cameras are carefully chosen~\cite{niemeyer2022regnerf, moreau2022lens}. Similarly, Manhattan SDF~\cite{guo2022manhattan} enforces Manhattan-world assumption by supposing that all surfaces are aligned with a canonical base. Nevertheless, strong assumptions of surface normals hinder their applicability to more complex scenes. To overcome this limitation, we propose a new regularization method that does not rely on any surface normal prior nor any geometry signal for training, resulting in accurate geometry reconstruction in large outdoor scenes for autonomous driving applications.

\section{Method}
\begin{figure*}[ht]
    \centering
    \includegraphics[width=0.94\textwidth]{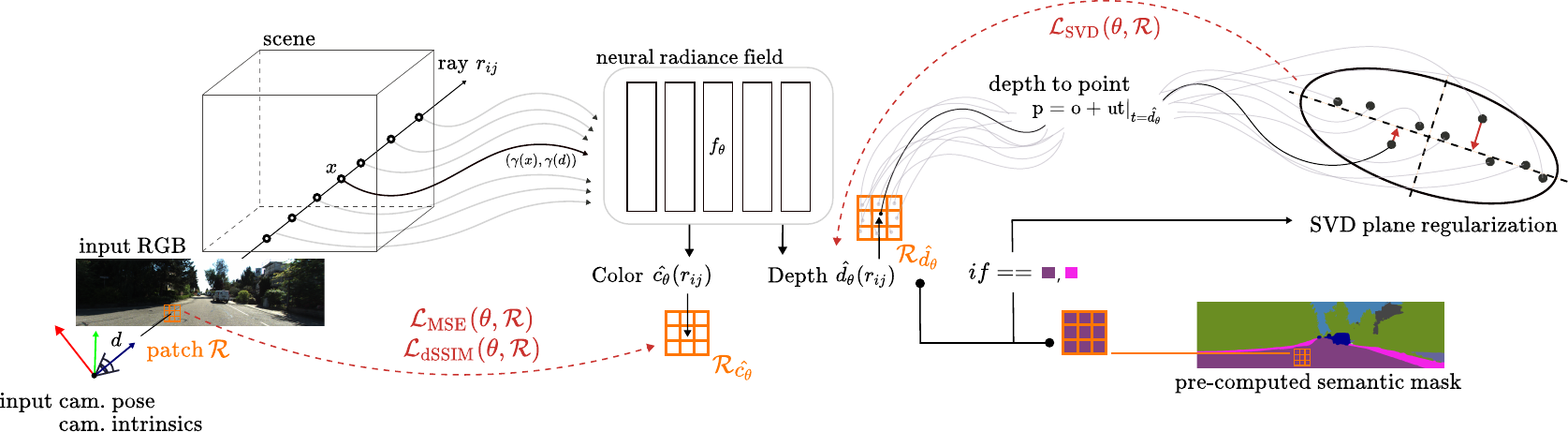} 
\caption{\textbf{Overview of PlaNeRF.} We improve upon NeRF training pipeline by appending a regularization term to the usual $L_2$ photometric loss. This is done by employing patch-based SSIM as a supervision signal enforced by $\mathcal{L}_{\text{dSSIM}}$ (Sec. \ref{sec:dssim_loss}) applied on image patches $\mathcal{R}_{\hat{c_{\theta}}}$. We additionally introduce a new regularization term $\mathcal{L}_{\text{SVD}}$ guided by semantic masks to regularize depth patches $\mathcal{R}_{\hat{d_{\theta}}}$ containing flat geometry to preserve sharp details at non-flat areas.}
    \label{fig:method_overview}
\end{figure*}

In this section, we discuss the details of our proposed novel regularization strategy to improve NeRF geometry for large outdoor scenes.
We propose a patch-based regularization method that constrains the geometry learnt by NeRF \textit{without} any prior 3D supervision. Specifically, we introduce a new plane regularization method on semantically selected classes that does not rely on surface normal priors based on singular value decomposition (see Section~\ref{sec:SVD_plane_regu}), and propose the use of the RGB patch SSIM for proper geometry initialization (see Section~\ref{sec:dssim_loss}). An overview of our method, coined PlaNeRF is depicted in Fig.~\ref{fig:method_overview}.

\subsection{Background}
\label{subsec:background}
\textbf{Neural Radiance Fields:}
a radiance field is a continuous function $f$ mapping a position and direction pair in 3D space $(\mathbf{x},\mathbf{u)} \in \mathbb{R}^3 \times \mathbb{S}^2$ to a volume density $\sigma \in \mathbb{R}^+$ and a color value $\mathbf{c} \in [0,1]^3$. \citet{mildenhall2021nerf} model this function with a multi layer perceptron (MLP) whose weights $\theta$ are optimized to reconstruct a 3D scene given a set of posed images during training.
\begin{equation}
\begin{split}
    f_{\theta}: \mathbb{R}^3 \times \mathbb{S}^2 \rightarrow [0,1]^3 \times \mathbb{R}^+\\
    (\gamma(\mathbf{x}), \gamma(\mathbf{u})) \rightarrow (\mathbf{c}, \sigma),
\end{split}
\end{equation}
where $\gamma$ represent a predefined spatial feature network or positional encoding.

\noindent\textbf{Volumetric rendering:}
given a neural radiance field, a pixel is rendered by casting a ray $\mathbf{r}(t) = \mathbf{o} + t\mathbf{u}$ from the camera center $\mathbf{o}$ through the pixel along direction $\mathbf{u}$.
For given rendering boundaries $t_n$ and $t_f$, the predicted color $\hat{c_{\theta}}(\mathbf{r})$ and depth $\hat{d_{\theta}}(\mathbf{r})$ of the ray are estimated by :
\begin{equation}
\begin{split}
    \hat{c_{\theta}}(\mathbf{r}) &=  \int_{t_n}^{t_f} T(t) \sigma \left( \mathbf{r}(t) \right) \mathbf{c_{\theta}}(\mathbf{r}(t), \mathbf{u}) \ dt, \\
    \hat{d_{\theta}}(\mathbf{r}) &= \int_{t_n}^{t_f} T(t) \sigma\left(\mathbf{r}(t)\right)t \ dt,\\
    \text{where} \ \ &T(t) = \exp \left( \int_{tn}^{tf} \sigma(\mathbf{r}(s)) ds \right)
\end{split}
\end{equation}

In RegNeRF, \citet{niemeyer2022regnerf} introduced a piece-wise smooth hypothesis for NeRF by proposing a depth-smoothness regularization term which aims to minimise the depth difference between adjacent pixels in a given image sub-patch:
\begin{equation}
    \begin{split}
        \mathcal{L_{\text{DS}}}(\theta, \mathcal{R})
        = \sum \limits_{\mathbf{r}\in \mathcal{R}} \sum\limits_{i,j = 1}^{\mathcal{S}_{\text{patch}}-1} & \left( \hat{d_{\theta}}(\mathbf{r}_{i,j}) - \hat{d_{\theta}}(\mathbf{r}_{i+1,j}) \right )^2 \\
        + & \left(\hat{d_{\theta}}(\mathbf{r}_{i,j}) - \hat{d_{\theta}}(\mathbf{r}_{i,j+1}) \right )^2,
    \end{split}  
\end{equation}
where $\mathcal{R}$ indicates the set of rays sampled for a patch of pixels, $\mathbf{r}_{ij}$ indicates the ray through pixel $(i,j)$, and $\mathcal{S}_{\text{patch}}$ is the size of the rendered patch.

As it can be deduced from the equation, such formulation enforces similar depth values on neighbouring pixels. In other words, depth-smoothing attempts to regularize the patch points to form a plane orthogonal to the center ray. Nevertheless, such strong prior on the surface normal limits its application to larger real-world scenes given the need of carefully-chosen virtual camera poses which can be time-consuming.

\subsection{Plane Regularization with SVD}\label{sec:SVD_plane_regu} 
We inspire from the patch-based optimization and piece-wise planar assumption introduced in RegNeRF~\cite{niemeyer2022regnerf}, but instead of regularizing the patch towards a plane orthogonal to the rays, we choose to target the least-squares plane approximation of the patch. This has the advantage of enabling the supervision signal to converge to the more adequate normal for each patch in an unsupervised manner.

Consider $\mathcal{R}$, the set of rays for a given patch, and its predicted point cloud $\mathcal{P} = \left\{ \mathbf{p}_r = \mathbf{o} + \hat{d}_{\theta_r}\mathbf{u_r} \right\}_{\mathbf{r} \in \mathcal{R}}$, where $\hat{d}_{\theta_r}$ and $\mathbf{u_r}$ are the rendered depth and direction of the ray $\mathbf{r}$ respectively. Its least-squares plane defined by a point and a normal unitary vector $(\hat{\mathbf{p_c}} , \hat{\mathbf{n}})$ is the solution of the optimization problem:
\begin{equation}
    \min \limits_{\mathbf{p_c}, \mathbf{n}} \sum \limits_{\mathbf{r} \in \mathcal{R}} \left((\mathbf{p}_r - \mathbf{p}_c) \cdot \mathbf{n} \right)^2.
    \label{equ: plan_regression}
\end{equation}

From the equation it can be deduced that a valid candidate for $\textbf{p}_c$ is the barycenter of the point cloud $$\hat{\textbf{p}}_c = \frac{1}{N}\sum \limits_{\mathbf{r} \in \mathcal{R}}  \textbf{p}_r$$

Let $\mathbf{A} = \begin{bmatrix}
\mathbf{p_0} - \hat{\textbf{p}}_c & \mathbf{p_1} - \hat{\textbf{p}}_c & \cdots & \mathbf{p_N} - \hat{\textbf{p}}_c
\end{bmatrix}^T$

be the $N \times 3$ matrix of the difference between each point and the barycenter, where $N = \mathcal{S}_{patch}^2$ is the number of points in the patch. The regression problem expressed in (\ref{equ: plan_regression}) can hence be solved by singular value decomposition, $\mathbf{A} = \mathbf{U}  \mathbf{S} \mathbf{V}^T$, where $\textbf{U}$ and $\textbf{V}$ are such that $\textbf{U}^T \textbf{U} =\textbf{V}^T \textbf{V}=\textbf{I}_3$ and $\textbf{S}$ is a square diagonal matrix:
\begin{align}
    \begin{split}
    \sum \limits_{\mathbf{r}_i \in \mathcal{R}_r} \left((\mathbf{p}_r - \hat{\textbf{p}_c}) \cdot \mathbf{n} \right)^2 &= \textbf{n}^T \textbf{A}^T \textbf{A n}\\
    =  \textbf{n}^T \textbf{VSU}^T\textbf{USV}^T\textbf{n}
    &=  \textbf{n}^T \textbf{VSSV}^T\textbf{n}\\
    &= \sum_i \sigma_i^2 ||\textbf{v}_i \cdot \textbf{n}||^2
    \label{eq: svd_simplification}
    \end{split}
\end{align}
where $\sigma_i$ is the $i^{th}$ singular value, arranged in descending order, $\textbf{v}_i$ is the $i^{th}$ column of $\textbf{V}$ and $n \in \mathbb{R}^3$.%

The solution to such equation can then obtained with $\textbf{n} = \textbf{v}_3$. That is, the least squares plan is defined by the barycenter of $\mathcal{P}_r$ and the right singular vector corresponding to the smallest singular value. The smaller $\sigma_3$, the closer the points are to their least squares plane.

Regularizing the NeRF rendered points to this plane is then exactly equivalent to minimizing the smallest singular value of $A$. Furthermore, the differentiable property of the SVD~\cite{townsend2016svddiff} ensures compatibility of this method with current automatic differentiation frameworks. 

Finally, we formulate our plane SVD regularization loss as:
\begin{equation}
 \mathcal{L_{\text{SVD}}}(\theta, \mathcal{R}) = \sigma_3(\theta, \mathcal{R}) 
\end{equation}

\subsection{Initialization of NeRF Geometry}\label{sec:dssim_loss}
Since we do not enforce any normal prior, our regularization relies on the assumption that the NeRF geometry is correctly initialized. Hence, it is crucial to guide the NeRF geometry to a good estimate before applying our SVD loss. 

To achieve this, we propose to supervise not only the pixel-wise consistency, but also the 2D local structure of the image. The 2D local structure supervision is computationally efficient for patch-based optimization and aids 3D estimation, as shown by~\citet{godard2017unsupervised}, who use dSSIM as a training loss for monocular self-supervised depth prediction. Inspired by that, we propose the use of patch dSSIM, to initialize the geometry of the learnt volumetric representation.

For given predicted RGB patch $\mathcal{X}_{\theta, \mathcal{R}}$ and its ground truth $\mathcal{Y}_\mathcal{R}$, we formulate the dSSIM supervision term as:
\begin{equation}
\begin{split}
    \mathcal{L}_{\text{dSSIM}}(\theta, \mathcal{R})
    & = \mathcal{L}_{\text{dSSIM}}( \mathcal{X}_{\theta, \mathcal{R}}, \mathcal{Y}_{ \mathcal{R}})\\
    & = \frac{1 - \text{SSIM}(\mathcal{X}_{\theta, \mathcal{R}}, \mathcal{Y}_{\mathcal{R}})}{2},
\end{split}
\end{equation}
where $\mathcal{R}$ is the set of rays for the RGB patch and the SSIM is a popular metric used in computer vision to evaluate the similarity between two images:

 \begin{equation}
     \text{SSIM}(x,y) = \frac{(2\mu_x\mu_y +c_1)(\sigma_{xy} + c_2)}{(\mu_x^2 + \mu_y^2 + c_1)(\sigma_x^2 + \sigma_y^2 +c_2)},
 \end{equation}
 
\noindent where $\mu_x, \mu_y$ are the pixel sample mean of image $x$ and $y$ respectively, $\sigma_x^2$ and $\sigma_y^2$ are the variance and $\sigma_{xy}$ is the covariance of $x$ and $y$. 
 
We have found in our experiments that by supervising not only the image color (\textit{i.e.} by minimizing the $L_2$ distance), but also the 2D structure can help the NeRF to improve its initial geometry, especially for low texture surfaces. We invite the reader to refer to Section \ref{sec:ablation} for an insight into how SSIM supervision impacts the final geometry accuracy.
 
\subsection{Training Strategy}
\label{subsec:training_strategy}
\textbf{Total Loss:} Our method is trained end-to-end from scratch by optimizing at each iteration for a given patch $\mathcal{R}$ the loss: 

 \begin{equation}
    \begin{split}
        \mathcal{L}(\theta, \mathcal{R}) =
        \mathcal{L}_{\text{MSE}} (\theta, \mathcal{R}) &+ \lambda_0 \mathcal{L}_{\text{dSSIM}} (\theta, \mathcal{R}) \\&+  \lambda_1 \mathcal{L}_{\text{SVD}} (\theta, \mathcal{R}),
    \end{split}
 \end{equation}

\noindent where $\mathcal{L}_{\text{MSE}}$ is the standard $L_2$ photometric loss used by NeRF, and $\lambda_i, i\in{0, 1}$ are weight coefficients used for each regularization term.

\noindent \textbf{Semantic masking:} 

PlaNeRF formulation requires a single plane segmentation prior in order to be effective. For instance, ``road", ``sidewalk", or ``traffic lanes" make such an assumption because multiple pixels labeled with the same class belong to a single 3D plane. This assumption is incorrect for other planar semantic classes, such as ``wall" or ``building", because many planes could be grouped into the same label. Therefore, we use semantic segmentation guidance to only regularize the aforementioned classes. Notice that these classes are predominant in driving scenarios, which makes PlaNeRF well-suited for such scenery.

\noindent \textbf{Loss scheduling:} to avoid inaccurate plane estimation due to uninitialized geometry, we have also delayed the plane regularization term by one epoch ($\lambda_1=0$ during the first epoch).

\section{Result and Experiments}
\subsection{Experimental Setup}
\textbf{Datasets:}
our method is evaluated on the KITTI-360 dataset~\cite{liao2022kitti}, which consists of five urban outdoor scenes between 50-100 meters long. Each scene has complex outdoor elements, including low texture surfaces and varying illumination conditions. 
Our network takes as input RGB images and camera poses estimated by using COLMAP~\cite{schonberger2016colmap}. Segmentation maps pre-computed from SegFormer~\cite{xie2021segformer} are considered to apply $\mathcal{L}_{\text{SVD}}$ over \textit{a-priori} flat-surface classes on the scene.

\noindent\textbf{Base architecture:}
we use a combination of existing methods as our base architecture, with hash encoding \cite{muller2022instant} and proposal networks~\cite{barron2022mip360}, similar to the Nerfacto model of nerfstudio\footnote{https://docs.nerf.studio/en/latest/nerfology/methods/nerfacto.html}. See appendix for implementation details of both proposal and NeRF network. We refer to this base architecture as base-NeRF. \textit{On top of our base-NeRF}, we apply the proposed patch-based geometry regularization method with dSSIM and SVD plane regularization, which we refer to as PlaNeRF.

\noindent\textbf{Implementation details:}
for all experiments, we use the Adam optimizer~\cite{kingma2014adam} and apply a cosine learning rate decay from 1e-2 to 1e-4 on a single NVIDIA GeForce RTX 3090 GPU. We set the batch size to 128 patches and train for 100 epochs with early stopping. We use ray patches of size  $\mathcal{S}_{\text{patch}} = 20$ to construct our training batches for both dSSIM and SVD plane regularization and we set $\lambda_0 = 0.1$ and $\lambda_1 = 0.01$. Training took approximately 1 to 4 hours by scene, depending on the convergence speed.
During training, the SVD plane regularization was only applied to concerned semantic classes given the strong flat surface prior, including road, traffic lanes and sidewalk. Following common practices~\cite{liao2022kitti}, we evaluate our method on KITTI-360 by using the official train/val split, where two front cameras are used for training with 50\% frame dropout rate and validation views are considered from the left front camera only.

\subsection{Results on Novel View Synthesis}

\begin{table}
    \centering
    \small
    \begin{tabular}{lccc}
        \hline
        Method   & PSNR~$\uparrow$ & SSIM~$\uparrow$   & LPIPS~$\downarrow$ \\ \hline
        AIR-SURF~\cite{wu2023mars}      & \textbf{23.09}           & \textbf{0.857}              & \textbf{0.174}              \\
        NeRF~\cite{mildenhall2021nerf}    & 21.18           & 0.779              & 0.343              \\
        PBNR~\cite{kopanas2021bpnr}       & 19.91           & 0.811  & \underline{0.191}     \\
        Mip-NeRF~\cite{barron2021mip} & 21.54           & 0.778              & 0.365              \\
        PNF~\cite{kundu2022panoptic}      & 22.07           & 0.820              & 0.221              \\
        Nerflets~\cite{zhang2023nerflets} & 21.69           & -                 & -                  \\
        \hline
        \textbf{PlaNeRF (ours)}& \underline{22.90}  & \textbf{0.857}     & 0.202                  \\ \hline
    \end{tabular}
    \caption{Results on KITTI-360 NVS benchmark~\cite{liao2022kitti}. Best results shown in \textbf{bold} and second best \underline{underlined}.}
    \label{tab:kittiBenchmark}
\end{table}

\begin{figure*}
	\centering
	\centering
	\setlength{\tabcolsep}{0.008\linewidth}
	\renewcommand{\arraystretch}{0.8}
	\begin{tabular}{cc}
		Base-NeRF & PlaNeRF (ours) \\
		\includegraphics[width=0.9\columnwidth]{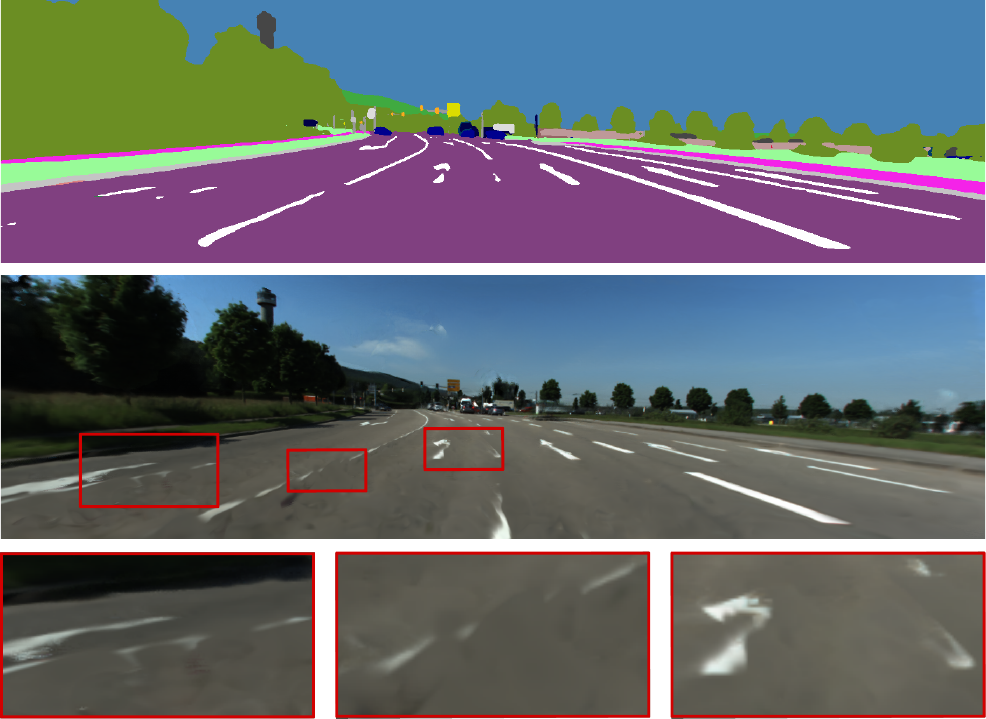} &
		\includegraphics[width=0.9\columnwidth]{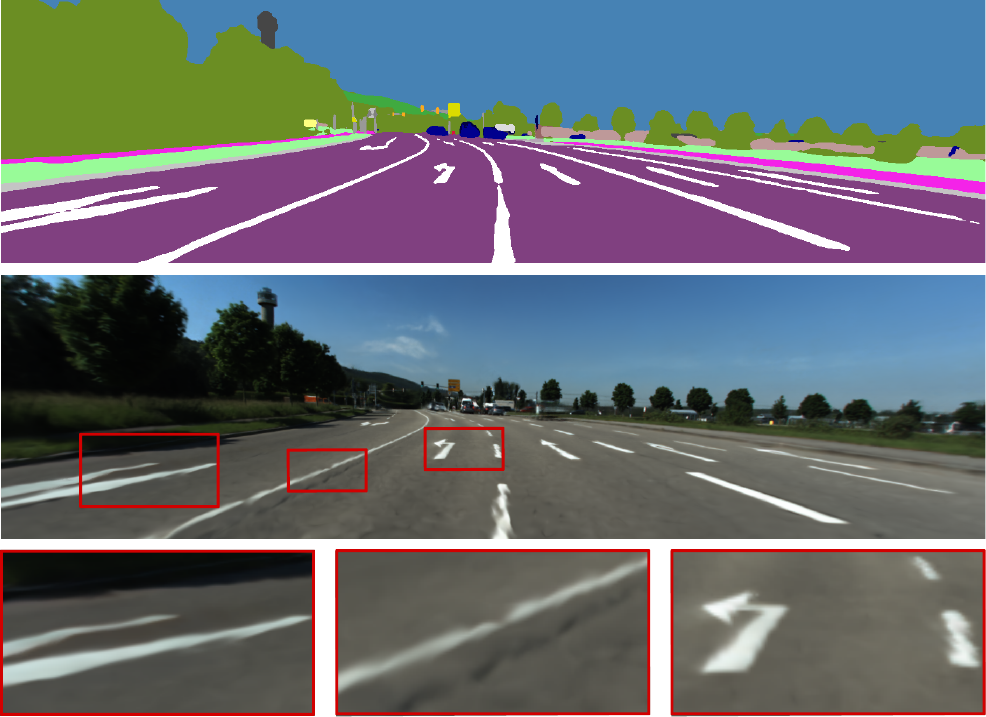} \\
  
	\end{tabular}
	\caption{\textbf{Qualitative evaluation of extrapolated NVS}. NeRF provides high quality images for novel views within the vicinity of training poses but rendering quality degrades when we extrapolate views due to incorrect 3D geometry (left). Our regularization method (right) leads to accurate 3D NeRF consistency without 3D supervision, as shown by rendered rgb images (bottom) and down-stream semantic segmentation predictions from SegFormer~\cite{xie2021segformer} (top).}
	\label{fig:extrapolation}
\end{figure*}
In the following, we report performance of our method against all published methods for outdoor NVS on the KITTI-360 benchmark\footnote{By the time of submission August 2023}: NeRF~\cite{mildenhall2021nerf}, PNBR ~\cite{kopanas2021bpnr}, Mip-NeRF~\cite{barron2021mip}, PNF~\cite{kundu2022panoptic}, AIR-SURF\cite{wu2023mars} and Nerflets~\cite{zhang2023nerflets}.
Table~\ref{tab:kittiBenchmark} reports conventional NVS metrics: PSNR, SSIM and LPIPS. Even though PlaNeRF is designed to improve scene geometry, our method achieves comparable image quality rendering to SOTA method. Furthermore, our geometry regularization strategy improves rendering quality (PSNR) without compromising image sharpness (SSIM) and achieves the best performance in SSIM. Compared to AIR-SURF and PNF, our proposal can only be applied on static scenes but do not rely on a scene-graph structure to describe the environment so it requires less pre-computation overhead and prior knowledge (e.g. 3D detection and instance segmentation) and can be trained much faster. At first glance, Nerflets is also a promising solution to describe outdoor scenes because of the spatially optimized small NeRFs used to fit the 3D structure of the environment. However, there is still a large gap in rendering performance when compared to our method.

\noindent\textbf{NVS from extrapolated poses:} the original setup of the KITTI-360 NVS benchmark is not well suited to compare geometry consistency, as interpolated novel views for evaluation are bounded between training ones. In order to showcase the extrapolation capability of our model, we setup another experiment where we generate novel views from a virtual camera executing a helix trajectory around the training poses shifted to the left by 1.9 meters. Notice that such setup can only be evaluated qualitatively.
Therefore, in Figure~\ref{fig:extrapolation}, we compare the NVS results before and after adding our regularization term. While base-NeRF fails to maintain the road lane details, PlaNeRF grants the model the ability to maintain photometric and geometry consistency at drastically extrapolated novel views, which can be seen by outputted continuous and clear lane details thanks to improved learnt geometry. Furthermore, we highlight semantic segmentation predictions on both outputs which clearly depicts more detailed road lanes and markings after applying our regularization term. We invite the reader to refer to our supplementary video for more qualitative evaluation.

\subsection{Results on Geometry Reconstruction}\label{sec:geometric_regu_comparison}

\begin{table}
    \small
    \centering
    \setlength{\tabcolsep}{0.016\linewidth}
    \begin{tabular}{cccccc}
        \hline
        Losses & CD~$\downarrow$ & $\text{P}_{\sigma}$~$\downarrow$ & PSNR~$\uparrow$ & SSIM~$\uparrow$ & LPIPS~$\downarrow$ \\ \hline
        $\mathcal{L}_{\text{dSSIM}}$ & 11.8 & 15.0 & \textbf{22.97} & \textbf{0.857} & \underline{0.206} \\
        $\mathcal{L}_{\text{dSSIM}} + \mathcal{L}_{\text{DS}}^*$~\cite{niemeyer2022regnerf}  & \underline{10.9} & 7.1 & 22.50 & 0.845 & 0.225 \\
        $\mathcal{L}_{\text{dSSIM}} + \mathcal{L}_{\text{Dist}}^*$~\cite{barron2022mip360} & 11.4 & \underline{6.8} & 22.86 & \underline{0.854} & 0.207 \\
        $\mathcal{L}_{\text{dSSIM}} + \mathcal{L}_{\text{Diff}}^*$~\cite{ehret2022diffnerf} & 12.8 & 9.5 & 22.75 & 0.850 & 0.224 \\ \hline
        PlaNeRF (ours) & \textbf{9.6} & \textbf{4.6} & \underline{22.90} & \textbf{0.857} & \textbf{0.205} \\ \hline
        {\scriptsize * Own implementation.}
    \end{tabular}
    \caption{Geometry accuracy on KITTI-360 benchmark of different SoTA regularization methods on top of base-NeRF and standard $\mathcal{L}_{\text{MSE}}$.
    Best results shown in \textbf{bold}, second best in 
    \underline{underlined}}
    \label{tab:methodCompare}
\end{table}

\begin{figure*}
	\centering
	\centering
	\setlength{\tabcolsep}{0.005\linewidth}
	\renewcommand{\arraystretch}{0.8}
	\begin{tabular}{ccc}
		$\mathcal{L}_{\text{Diff}}$~\cite{niemeyer2022regnerf} & $\mathcal{L}_{\text{DS}}$~\cite{ehret2022diffnerf} & PlaNeRF (ours) \\
		\includegraphics[width=0.64\columnwidth]{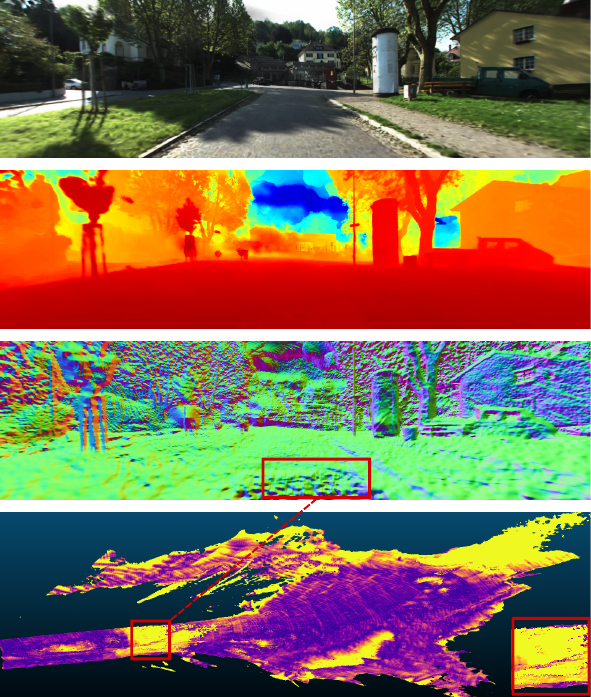} &
		\includegraphics[width=0.64\columnwidth]{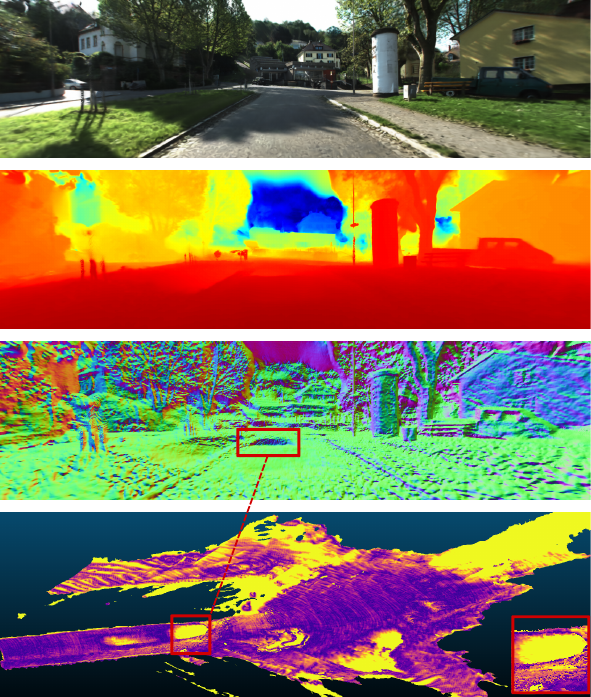} &
		\includegraphics[width=0.64\columnwidth]{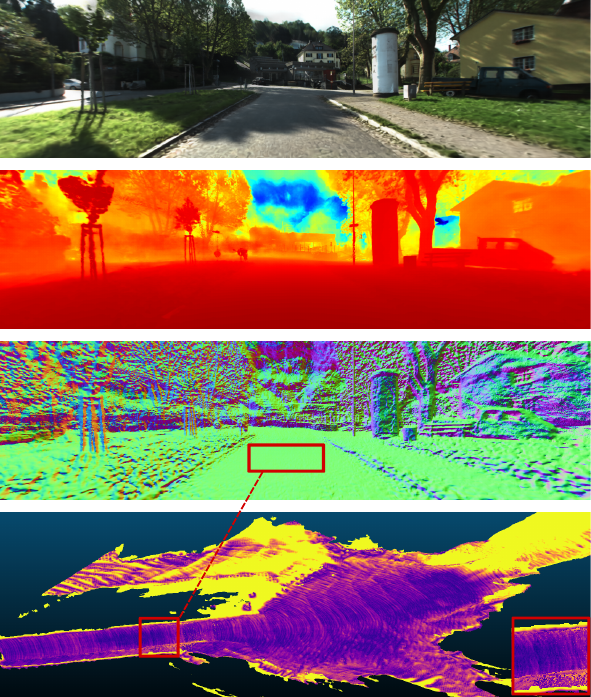} \\

        \multicolumn{3}{c}{\includegraphics[width=2.00\columnwidth]{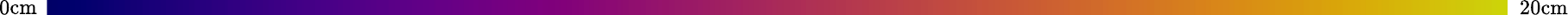}}
  
	\end{tabular}
 	\caption{\textbf{Geometric evaluation.} We show qualitative results of different regularization methods applied on top of base-NeRF. From top to bottom: rendered RGB on test poses, normal maps, depth maps and Poisson mesh coloured by distance to groundtruth 
    . We find both $\mathcal{L}_{\text{DS}}$ and $\mathcal{L}_{\text{Diff}}$ fail to reconstruct road surface while PlaNeRF provides the best accurate geometry on road and sidewalk.}
 
	\label{fig:method_compare}
\end{figure*}

To highlight the performance of our SVD plane regularization, we perform exhaustive evaluation and comparisons against other popular geometry regularization techniques, namely:

\begin{itemize}
    \item Depth Smoothness loss $\mathcal{L}_{\text{DS}}$ from~\cite{niemeyer2022regnerf},
    \item Depth Differential loss $\mathcal{L}_{\text{Diff}}$ from~\cite{ehret2022diffnerf},
    \item Distortion loss $\mathcal{L}_{\text{Dist}}$ from~\cite{barron2022mip360}.
\end{itemize}

As the aforementioned regularization have been implemented in different frameworks, we re-implement them on top of our base-NeRF to isolate the effectiveness of each solution. For fair comparison, the coefficient for each regularization method is carefully tuned on the KITTI-360 dataset and all the other losses ($\mathcal{L}_{\text{MSE}}$ and $\mathcal{L}_{\text{dSSIM}}$).

In addition to traditional NVS metrics for NeRF, we use two additional metrics to evaluate the geometry of the NeRF 3D reconstruction: the point cloud chamfer distance and the plane standard deviation. Although our methods do not require 3D supervision, both evaluation metrics require LiDAR point cloud as ground truth.

\noindent\textbf{Point Cloud Chamfer Distance (CD):}
we utilize the KITTI-360 LiDAR point cloud as a reference ground truth. We select LiDAR rays that terminates on semantic classes that should be flat and pass them to NeRF for depth estimation (see supplementary materials for more details).
Subsequently, we obtain a second LiDAR point cloud rendered by NeRF's prediction.
The mean chamfer distance (CD) between these two accumulated point clouds is then calculated as:
\begin{equation}
\begin{split}
    \text{CD}(X, Y) = &\frac{1}{2N} \sum \limits_{x \in X} \min \limits_{y \in Y} ||x-y||^2\\ 
    + &\frac{1}{2M} \sum \limits_{y \in Y} \min \limits_{x \in X} ||x-y||^2
\end{split}
\end{equation}
Where $N, M$ are the number of points in point cloud $X, Y$ respectively.

\noindent\textbf{Plane Standard Deviation ($\text{P}_{\sigma}$):}
to specifically evaluate the plane reconstruction, we calculate the Standard Deviation of NeRF predicted points along the estimated normal from LiDAR pointcloud, which we refer to as $\text{P}_{\sigma}$:

\begin{equation}
\begin{split}
    &\text{P}_{\sigma}(\mathcal{P}_X, \mathcal{P}_Y) = \sigma(\mathcal{P}_X \cdot \mathbf{n_{\mathcal{P}_Y}}),
\end{split}
\end{equation}
where $\sigma$ is the standard deviation and $\mathcal{P}_X, \mathcal{P}_Y$ represent point patches obtained from NeRF and LiDAR groundtruth respectively, and $\mathbf{n}$ is the surface normal estimated from the LiDAR point cloud patch $\mathcal{P}_Y$.
We argue that this metric is more robust to calibration offset and provides a more direct assessment of NeRF's ability to reconstruct planar surfaces.

\noindent\textbf{Results:} we report results of our base-NeRF trained with all previously described regularization losses in Table~\ref{tab:methodCompare}. From the metrics it can be deduced that PlaNeRF gives the best overall geometry on flat surface reconstruction (lowest CD and $\text{P}_{\sigma}$) while maintaining a good rendering quality (second best PSNR and pair best SSIM). Naturally, isolated $\mathcal{L}_{\text{dSSIM}}$ achieves the best rendering quality as concerned metrics are closer to the optimization objective. Although $\mathcal{L}_{\text{DS}}$ and $\mathcal{L}_{\text{Diff}}$ are effective to smooth geometry of object-centric scenes, we demonstrate from our experiments that it is less effective for recovering good structure in autonomous driving scenarios. 
This can be explained by the fact that 
different camera rays can hit the same surface from similar directions on autonomous driving setups, which might induce bias in normals estimation
(see Section~\ref{subsec:background}). Even if $\mathcal{L}_{\text{Dist}}$ only operate at a ray level (compared to patches of rays used in PlaNeRF), it helps the model to improve drastically the learnt geometry. However, our proposal outperforms all others in reconstructing flat surfaces thanks to the planar constraint induced by our formulation. Qualitative results are presented in Fig.~\ref{fig:method_compare}.  It can be observed that using $\mathcal{L}_{\text{DS}}$ and $\mathcal{L}_{\text{Diff}}$ regularization generates a noisy reconstruction of large surfaces such as the road plane, resulting in "bumpy" non-flat surface. In contrast, our proposed SVD plane regularization loss ($\mathcal{L}_{\text{SVD}}$) succeeds in recovering an accurate and smooth plane surface at such areas. We provide more visual comparison in the supplementary material.

\begin{table}[]
    \small
    \centering
    \setlength{\tabcolsep}{0.014\linewidth}
    \begin{tabular}{lcccccc}
        \hline
        & CD~$\downarrow$ & $\text{P}_{\sigma}$~$\downarrow$ & PSNR~$\uparrow$ & SSIM~$\uparrow$ & LPIPS~$\downarrow$ & iter/s~$\uparrow$\\ \hline
        PlaNeRF & \textbf{9.6} & \textbf{4.6} & \underline{22.90} & \textbf{0.857} & \textbf{0.205} & 5.6\\
        w/o $\mathcal{L}_{\text{SVD}}$ & 11.8 & 15.0 & \textbf{22.97} & \textbf{0.857} & \underline{0.206} & \textbf{10} \\ 
        w/o $\mathcal{L}_{\text{dSSIM}}$ & \underline{11.3} & \underline{6.7} & 22.69 & \underline{0.837} & 0.246 & 5.6 \\ \hline
    \end{tabular}
    \caption{Ablation of our contributions on the KITTI-360 benchmark.}
    \label{tab:ablation}
\end{table}

\subsection{Ablation Studies}
\label{sec:ablation}
\begin{figure}
	\centering
	\centering
	\setlength{\tabcolsep}{0.005\linewidth}
	\renewcommand{\arraystretch}{0.8}
	\begin{tabular}{cc}
		w/o $\mathcal{L}_{\text{SVD}}$ & PlaNeRF (ours) \\
		\includegraphics[width=0.48\columnwidth]{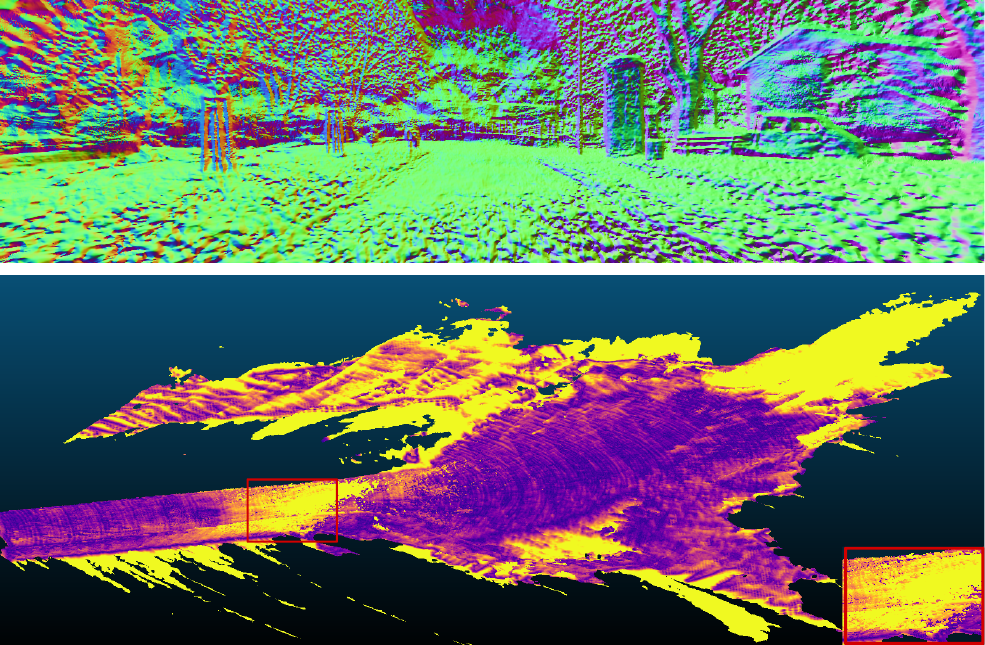} &
		\includegraphics[width=0.48\columnwidth]{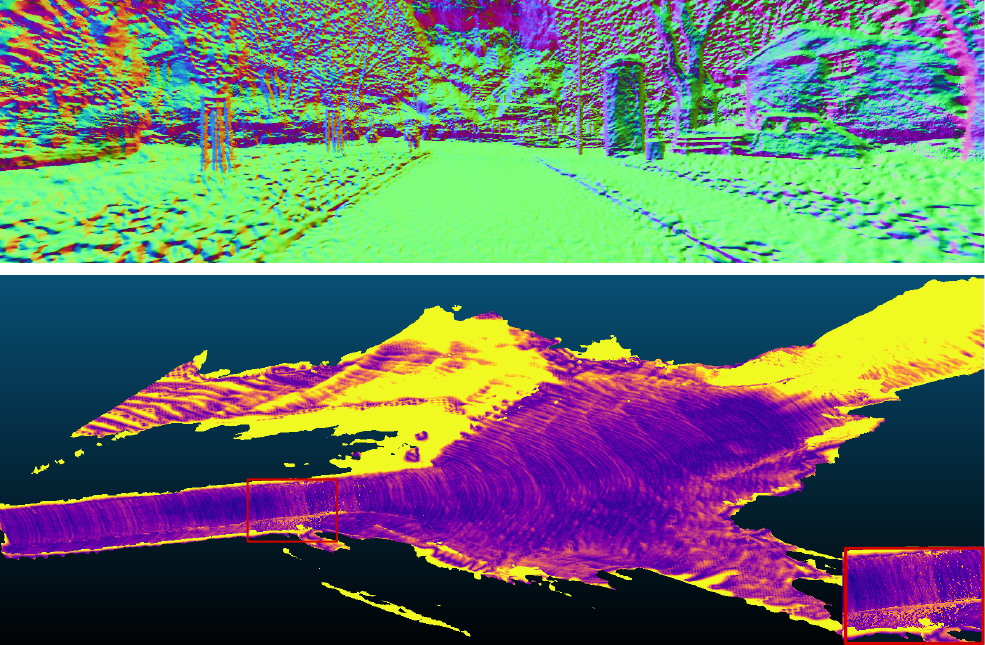}\\
      \multicolumn{2}{c}{\includegraphics[width=0.95\columnwidth]{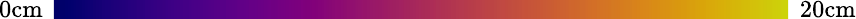}}
	\end{tabular}
 \caption{\textbf{Ablation study.} As shown by normal maps (top) and Poisson mesh colored by distance to groundtruth (bottom), 
 $\mathcal{L}_{\text{SVD}}$ regularization enhances geometry and produce smoother planar surfaces.}
	\label{fig:ablation}
\end{figure}
In Table~\ref{tab:ablation}, we perform ablation on different components of our regularization method, namely $\mathcal{L}_{\text{SVD}}$ and $\mathcal{L}_{\text{dSSIM}}$. We show the average geometry and rendering quality metrics for all five KITTI-360 scenes. From our experiments, we found out that enforcing RGB-patched $\mathcal{L}_{\text{dSSIM}}$ is very effective for NeRF geometry initialization and has a positive impact on rendering quality. In addition, adding patched $\mathcal{L}_{\text{SVD}}$ further improves the scene geometry and smooths planar surfaces by a large gap at the cost of small PSNR drop (we refer the reader to Fig. \ref{fig:ablation} and supplementary video for qualitative comparisons). 

Finally, in the last column of Table~\ref{tab:ablation}, we present the computation overhead of adding our proposed components. In particular, we observe that the inclusion of the $\mathcal{L}_{\text{dSSIM}}$ regularization has negligible impact on computation time. However, the inclusion of $\mathcal{L}_{\text{SVD}}$ leads to a 44\% decrease in training speed. The latter can be explained by our $\mathcal{L}_{\text{SVD}}$ implementation in \texttt{pytorch}, which becomes inefficient on GPU for small matrix sizes, requiring them to be passed to CPU.

\section{Conclusion}
We have presented PlaNeRF, a novel approach to regularize Neural Radiance Field (NeRF) without any 3D supervision. Our key insight is that for large driving scene reconstruction, NeRF usually fails to learn accurate geometry, especially at low texture area. 
To overcome this limitation, we propose a new plane regularization based on SVD without forcing any priors on the surface. Combined with SSIM supervision for proper geometry initialisation, our method accurately reconstructs low texture planar areas and improves scene geometry. Our experimental evaluation shows that our model outperforms current popular geometry regularization techniques on large driving scene reconstruction and achieves comparable results to SOTA method on KITTI-360 NVS benchmark.

A main limitation of our method is that relies on semantic segmentation masks (which might be wrong/inaccurate) and that can be applied only at \textit{a-priori} flat surfaces. In the future, we intend to find a semantically independent formulation to exploit the local shape of predicted geometry and determine weather the regularization should be applied in a fully unsupervised way.

{
    \small
    \bibliographystyle{ieeenat_fullname}
    \bibliography{main}
}

\clearpage
\appendix
\end{document}


\title{Supplementary Material for PlaNeRF: SVD Unsupervised 3D Plane Regularization for NeRF Large-Scale Scene Reconstruction}

\author{Fusang Wang\\
Huawei Technologies\\ 
Shanghai Jiao Tong University\\
{\tt\small }
\and
Arnaud Louys\\
Huawei Technologies\\
{\tt\small }
\and
Nathan Piasco\\
Huawei Technologies\\
{\tt\small }
\and
Moussab Bennehar\\
Huawei Technologies\\
{\tt\small }
\and
Luis Rold\~ao\\
Huawei Technologies\\
{\tt\small }
\and
Dzmitry Tsishkou\\
Huawei Technologies\\
{\tt\small }
}
\maketitle
\appendix
\section{Base-NeRF Implementation Details}
\begin{table*}[htbp]
\centering
\caption{Base-NeRF architecture. Hash Grid options: number of level/log2 hashmap size/base resolution/max resolution. MLP options: number of layers/hidden dimension.}
\label{tab:base_nerf}
\begin{tabular}{lcccc}
\hline
 Component & Positional embedding &  Directional embedding & Sigma MLP & Color MLP \\ \hline
 Proposal 1 & Hash Grid 5/17/16/64 & - & 1/16 & - \\
 Proposal 2 & Hash Grid 5/17/16/256 & - & 1/16 & - \\
 NeRF       & Hash Grid 5/19/16/2048 & Spherical harmonic & 1/64 & 2/64 \\
\hline
\end{tabular}
\end{table*}
We use an implicit scene representation similar to the \texttt{nerfacto} model of Nerfstudio\footnote{https://docs.nerf.studio/en/latest/nerfology/methods/nerfacto.html} open source framework. Details are shown in Table~\ref{tab:base_nerf}. We use 128 linear samples to query the first proposal network, respectively 96 and 48 samples using probability density sampling with second proposal respectively final NeRF model. At test time, we higher the number of sampled points for the final NeRF model to 192 in order get better rendering quality.

\section{Details on Semantic Guidance}
When implementing the SVD plane regularization, we specifically apply it only to semantic classes that are expected to have a flat surface, such as roads and sidewalks. Our goal is to enforce the regularization only when the patch represents a single plane in 3D space. In practical implementation, however, we encounter two challenging scenarios. First, it is possible for multiple semantic classes to exist within the same patch. In such cases, it's likely that there are multiple planes in that specific patch, and we should refrain from applying the regularization because the single plane assumption is violated. Second, multiple semantic classes may belong to the same plane. For example, a road and its lanes could be part of the same planar structure. In these cases, regularization should be applied even if multiple semantic classes are involved. To address these challenges, we group semantic classes that share the same plane into a single semantic group and apply the regularization only when a single semantic group is detected. In the future, we intend to generalize the SVD plane regularization to be independent of semantic labels and to be able to handle cases where multiple planes are detected within the same patch.

\section{KITTI-360 Extrapolation Sequence}
For the results in the teaser image and extrapolation NVS renderings, we select a scene from KITTI-360 sequence 0003, frame 687-717. The experimental setup for training and evaluation remains consistent with the default settings used for other KITTI-360 NVS sequences, where the two front cameras are used for training with 50\% frame dropout rate and validation views are considered from the left-front camera only.

\section{Geometry Evaluation Metrics}
\textbf{Details on LiDAR Semantic Filtering.}
To evaluate the Chamfer distance for specific semantic classes, we use the segmentation predictions generated by SegFormer to filter the ground truth LiDAR data provided by the KITTI-360 dataset. To obtain a point cloud that contains semantic classes of interest, we first identify the LiDAR points that fall within the camera's field of view. Then, we select from these points the rays that terminate on the desired semantic classes, such as roads and sidewalks. Finally, we feed these filtered LiDAR rays into the NeRF model to do depth renderings and predict the corresponding point cloud. For each test frame, we accumulate both the LiDAR ground truth and the NeRF-predicted point clouds to evaluate the quality of NeRF's geometry reconstruction. 

\noindent\textbf{Details on Plane Standard Deviation ($\mathbf{P}_{\sigma}$).}
To focus on the evaluation of plane reconstruction, we calculate the Standard Deviation of NeRF predicted points along the estimated normals from the LiDAR point cloud. Specifically, for a given semantic group representing a single plane, we extract the filtered LiDAR point cloud and the corresponding NeRF point cloud. We then split both point clouds into patches of size $3 \times 3$ square meters. For each pair of LiDAR and NeRF patches, we use the RANSAC algorithm to estimate the normal of the LiDAR patch. Then, we calculate the standard deviation of the patch along the estimated normal.

\section{By Scene Results}
Table~\ref{tab:resGeo} and Table~\ref{tab:resRender} present comprehensive results for both NVS and geometry evaluation in each scene of the KITTI-360 NVS benchmark. Our proposed method outperforms other approaches in terms of geometry accuracy, demonstrating superior performance in all five scenes. In addition, our method exhibits the least degradation in rendering quality compared to the competing methods, highlighting its effectiveness in preserving the visual fidelity of the reconstructed scenes.
\begin{table*}[htbp]
\centering
\caption{Per scene geometric accuracy on KITTI-360 benchmark of base-NeRF with different SoTA regularization methods. All methods are using standard $\mathcal{L}_{\text{MSE}}$ in addition to the regularization losses mentioned in the table. Best results shown in \textbf{bold}, second best shown in \underline{underlined}}
\label{tab:resGeo}
\begin{tabular}{cccccccc}
\hline
\textbf{Metrics} & Losses & Scene00 & Scene01 & Scene02 & Scene03 & Scene04 & Average \\ \hline
\multirow{5}{*}{CD~$\downarrow$} 
 & $\mathcal{L}_{\text{dSSIM}}$ & 11.2 & 10.2 & \underline{11.3} & 11.2 & 15.2 & 11.8 \\
 & $\mathcal{L}_{\text{dSSIM}} + \mathcal{L}_{\text{DS}}^*$ & \underline{9.4} & \underline{8.8} & 12.7 & \underline{10.2} & \underline{13.5} & \underline{10.9} \\
 & $\mathcal{L}_{\text{dSSIM}} + \mathcal{L}_{\text{Dist}}^*$ & 9.8 & 9.5 & 12.9 & 11.3 & 13.6 & 11.4 \\
 & $\mathcal{L}_{\text{dSSIM}} + \mathcal{L}_{\text{Diff}}^*$ & 10.7 & 12.1 & 15.1 & 11.0 & 15.1 & 12.8 \\ \cline{2-8} 
 & PlaNeRF (Ours) & \textbf{8.1} & \textbf{7.6} & \textbf{11.0} & \textbf{9.5} & \textbf{11.9} & \textbf{9.6} \\ \hline
\multirow{5}{*}{Plane std~$\downarrow$} 
 & $\mathcal{L}_{\text{dSSIM}}$ & 13.1 & 7.9 & 16.5 & 18.7 & 18.9 & 15.0 \\
 & $\mathcal{L}_{\text{dSSIM}} + \mathcal{L}_{\text{DS}}^*$& \underline{5.7} & 6.3 & 7.1 & \underline{6.4} & 9.8 & 7.1 \\
 & $\mathcal{L}_{\text{dSSIM}} + \mathcal{L}_{\text{Dist}}^*$ & 6.5 & \underline{6.0} & \underline{6.4} & 7.2 & \underline{7.9} & \underline{6.8} \\
 & $\mathcal{L}_{\text{dSSIM}} + \mathcal{L}_{\text{Diff}}^*$ & 6.6 & 10.4 & 12.4 & 7.9 & 10.1 & 9.5 \\ \cline{2-8} 
 & PlaNeRF(Ours) & \textbf{4.5} & \textbf{3.9} & \textbf{3.8} & \textbf{4.9} & \textbf{6.0} & \textbf{4.6} \\ \hline
 
 {\scriptsize * Own implementation.}
\end{tabular}
\end{table*}
\begin{table*}[htbp]
\centering
\caption{Per scene rendering results on KITTI-360 benchmark of base-NeRF with different SoTA regularization methods. }
\label{tab:resRender}
\begin{tabular}{cccccccc}
\hline
\multicolumn{1}{l}{\textbf{Metrics}} & Reg method & Scene00 & Scene01 & Scene02 & Scene03 & Scene04 & Average \\ \hline
\multirow{5}{*}{PSNR~$\uparrow$} 
 & $\mathcal{L}_{\text{dSSIM}}$ & \textbf{23.53} & \textbf{22.30} & \textbf{22.91} & \textbf{22.55} & \textbf{23.57} & \textbf{22.97} \\
 & $\mathcal{L}_{\text{dSSIM}} + \mathcal{L}_{\text{DS}}^*$ & 23.27 & 21.69 & 22.41 & 22.31 & 22.82 & 22.50 \\
 & $\mathcal{L}_{\text{dSSIM}} + \mathcal{L}_{\text{Dist}}^*$ & \underline{23.42} & 22.19 & 22.78 & 22.44 & 23.49 & 22.86 \\
 & $\mathcal{L}_{\text{dSSIM}} + \mathcal{L}_{\text{Diff}}^*$ & 23.29 & 22.13 & 22.72 & 22.27 & 23.35 & 22.75 \\ \cline{2-8} 
 & PlaNeRF (Ours) & \underline{23.42} & \underline{22.23} & \underline{22.79} & \underline{22.50} & \underline{23.54} & \underline{22.90} \\ \hline
\multirow{5}{*}{SSIM~$\uparrow$} 
 & $\mathcal{L}_{\text{dSSIM}}$ & \textbf{0.887} & \underline{0.808} & \textbf{0.855} & \textbf{0.867} & \underline{0.869} & \textbf{0.857} \\
 & $\mathcal{L}_{\text{dSSIM}} + \mathcal{L}_{\text{DS}}^*$              & 0.878 & 0.786 & 0.843 & 0.863 & 0.855 & 0.845 \\
     & $\mathcal{L}_{\text{dSSIM}} + \mathcal{L}_{\text{Dist}}^*$               & 0.884 & 0.803 & \underline{0.853} & \underline{0.865} & 0.866 & \underline{0.854} \\
 & $\mathcal{L}_{\text{dSSIM}} + \mathcal{L}_{\text{Diff}}^*$               & 0.881 & 0.799 & 0.849 & 0.859 & 0.863 & 0.850 \\ \cline{2-8} 
 & PlaNeRF (Ours)                   & \underline{0.886} & \textbf{0.809} & 0.851 & \textbf{0.867} & \textbf{0.870} & \textbf{0.857} \\ \hline
 \multirow{5}{*}{LPIPS~$\downarrow$} 
 & $\mathcal{L}_{\text{dSSIM}}$ & \textbf{0.177} & 0.237 & \textbf{0.205} & \underline{0.210} & \underline{0.200} & \underline{0.206} \\
 & $\mathcal{L}_{\text{dSSIM}} + \mathcal{L}_{\text{DS}}^*$ & 0.206 & 0.259 & 0.221 & 0.213 & 0.224 & 0.225 \\
 & $\mathcal{L}_{\text{dSSIM}} + \mathcal{L}_{\text{Dist}}^*$ & \underline{0.178} & \underline{0.233} & \underline{0.207} & 0.212 & 0.205 & 0.207 \\
 & $\mathcal{L}_{\text{dSSIM}} + \mathcal{L}_{\text{Diff}}^*$ & 0.194 & 0.257 & 0.224 & 0.224 & 0.220 & 0.224 \\ \cline{2-8} 
 & PlaNeRF (Ours)                   & 0.181 & \textbf{0.232} & \underline{0.207} & \textbf{0.207} & \textbf{0.199} & \textbf{0.205} \\ \hline
 {\scriptsize * Own implementation.}
\end{tabular}
\end{table*}
\begin{figure*}
	\centering
	\setlength{\tabcolsep}{0.005\linewidth}
	\renewcommand{\arraystretch}{0.8}
 	\begin{tabular}{ccc}
         PlaNeRF pointcloud & PlaNeRF mesh & LiDAR mesh (groundtruth) \\
		\includegraphics[width=0.65\columnwidth ]{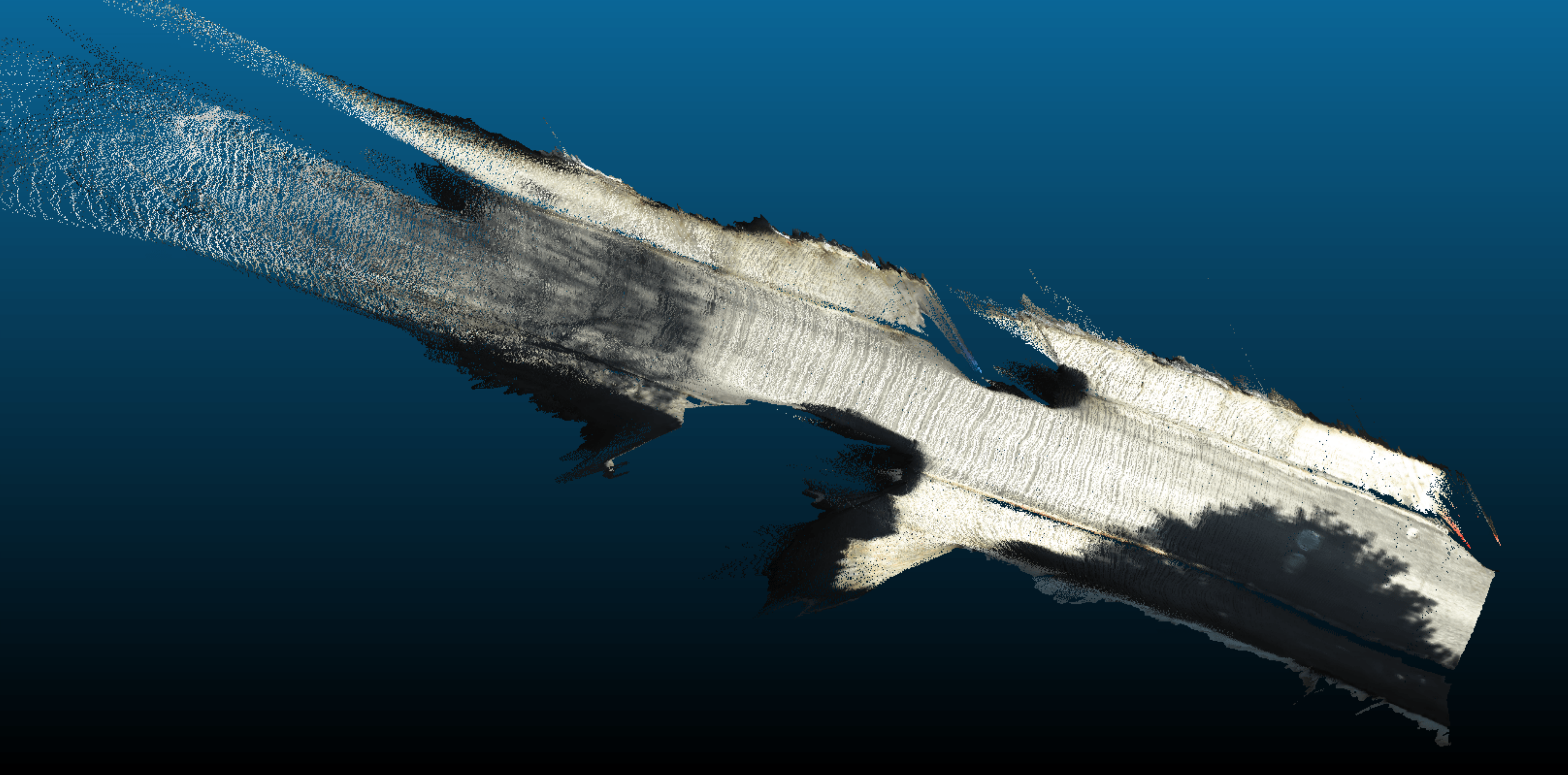} &
		\includegraphics[width=0.65\columnwidth ]{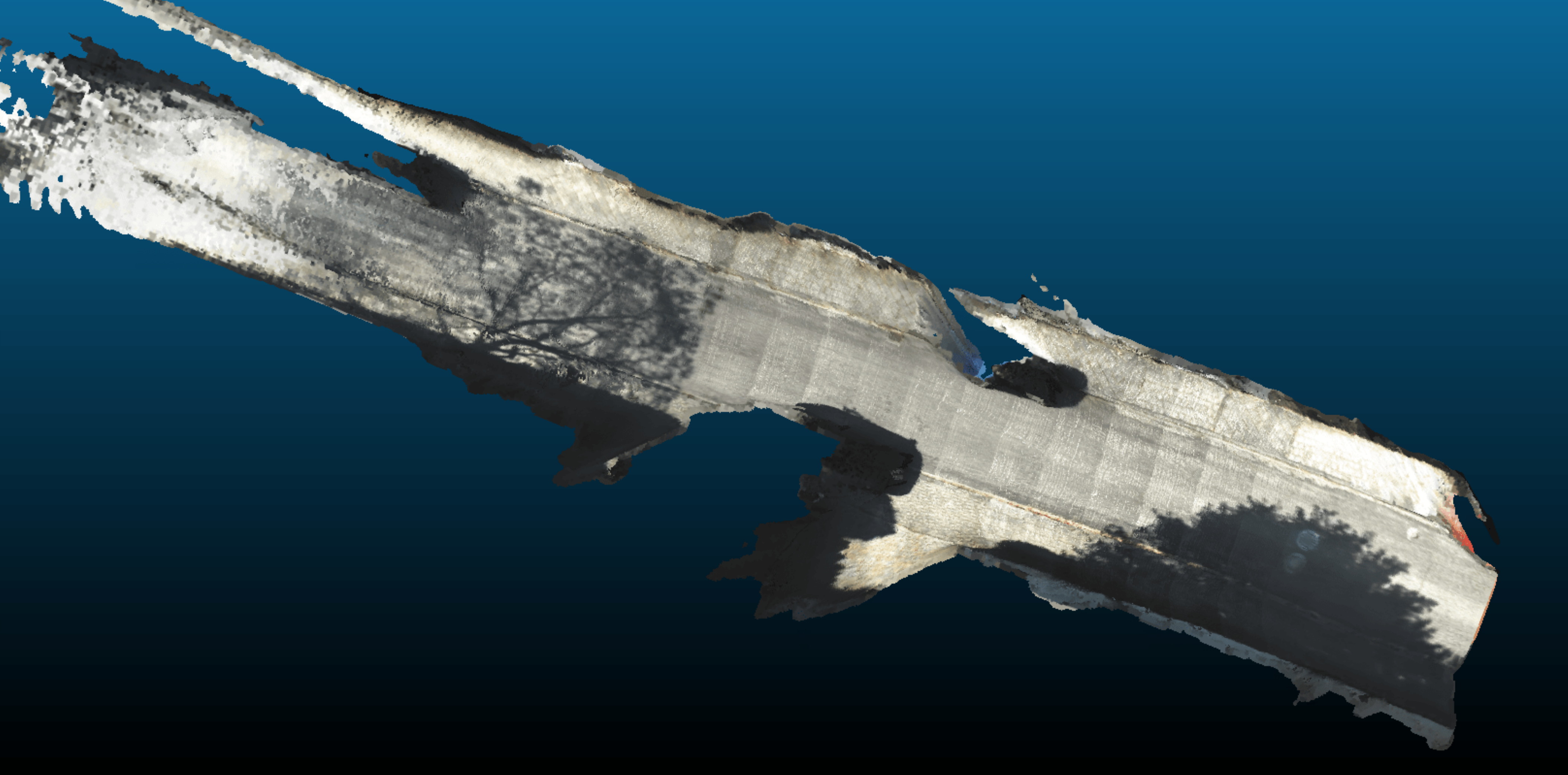} &
		\includegraphics[width=0.65\columnwidth ]{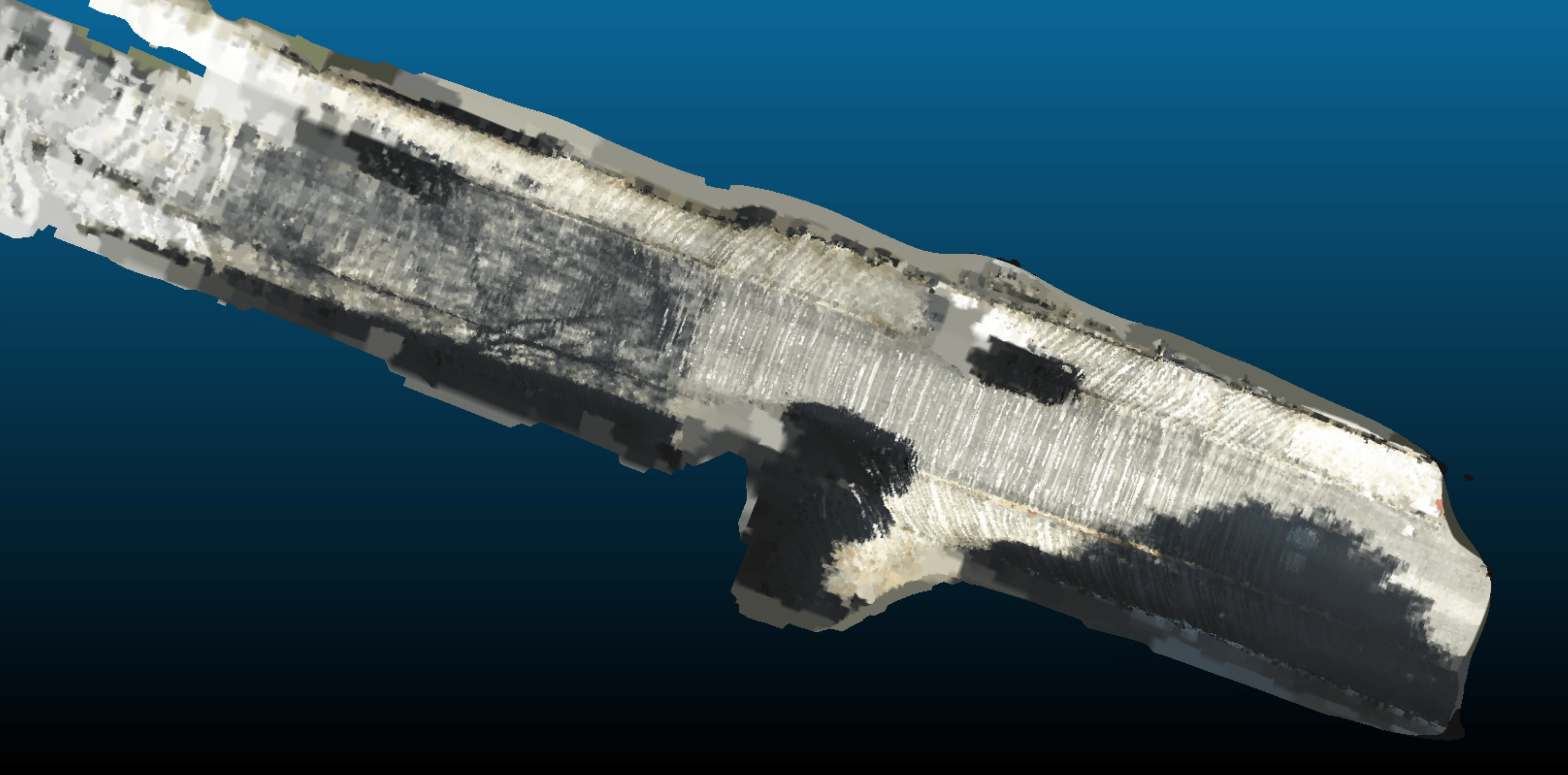} \\

		\includegraphics[width=0.65\columnwidth ]{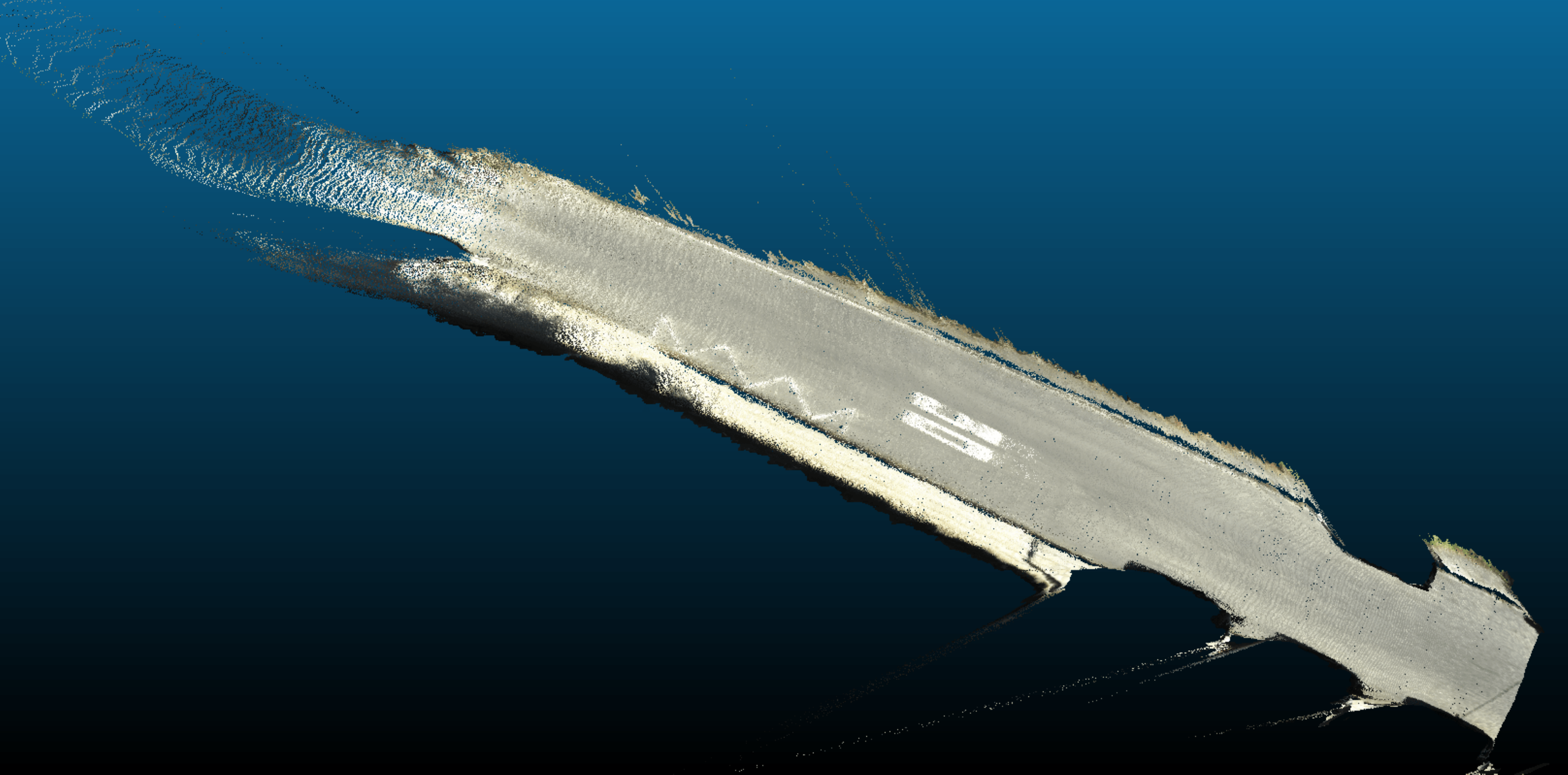} &
		\includegraphics[width=0.65\columnwidth ]{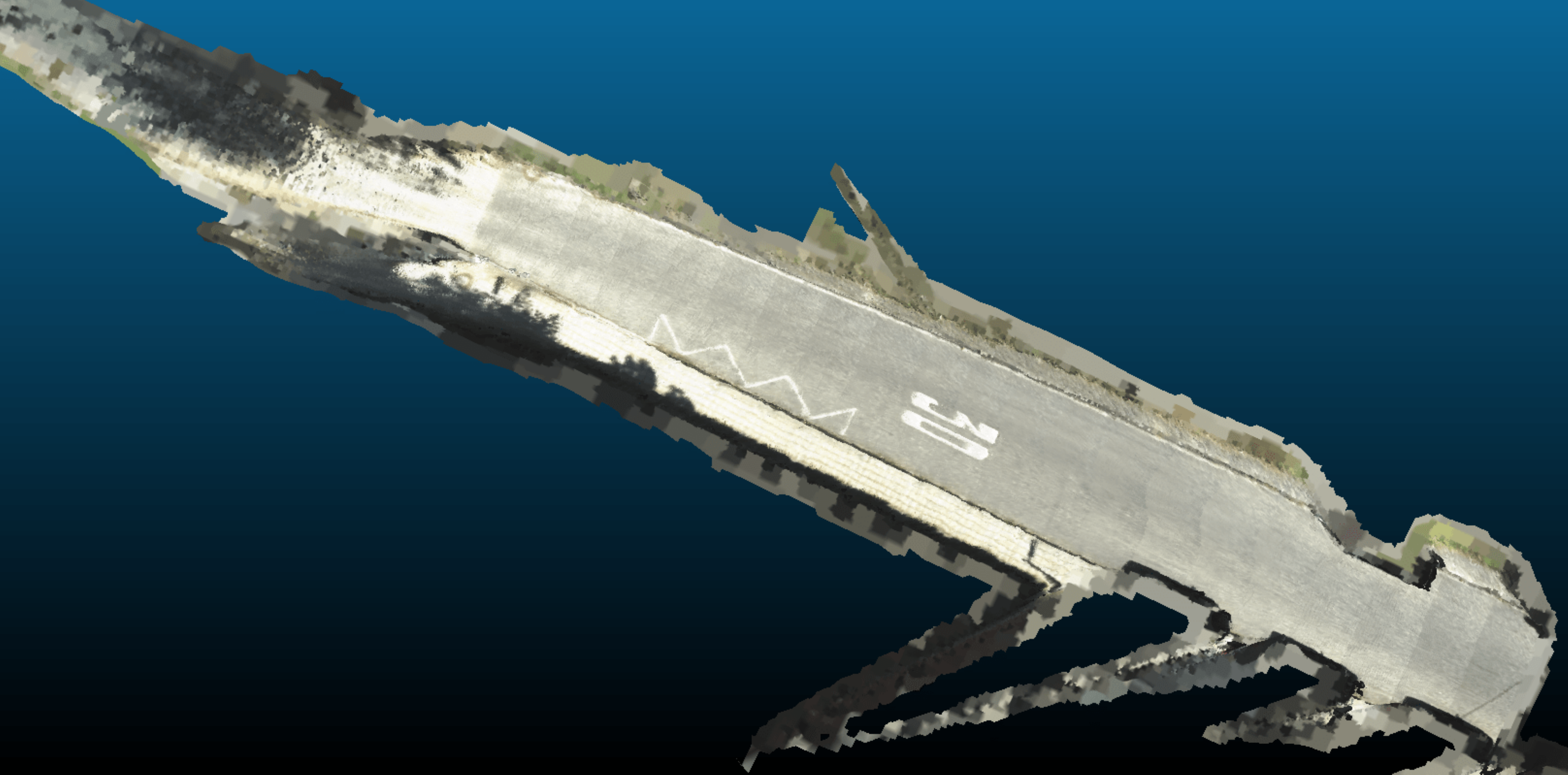} &
		\includegraphics[width=0.65\columnwidth ]{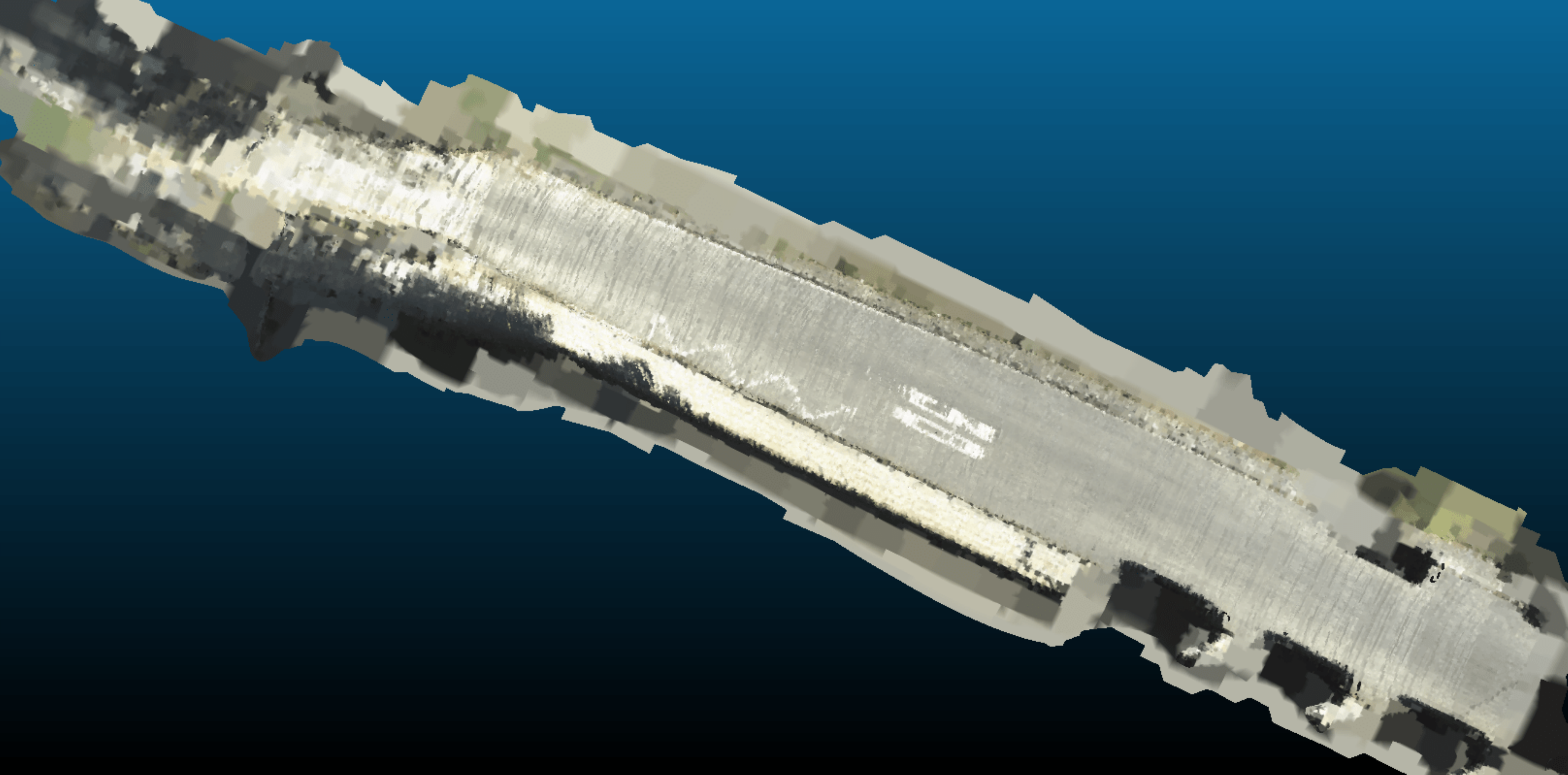} \\

		\includegraphics[width=0.65\columnwidth ]{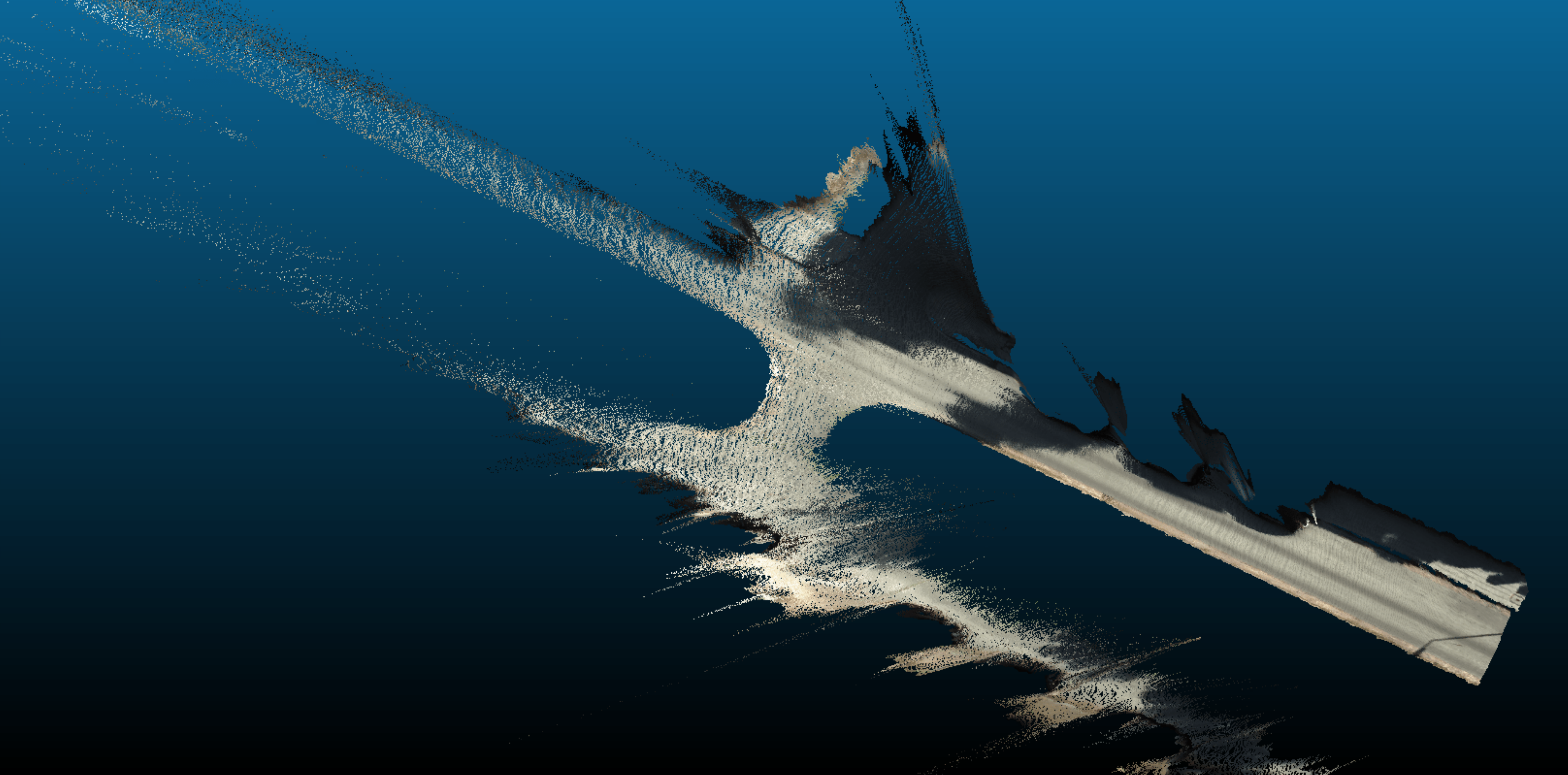} &
		\includegraphics[width=0.65\columnwidth ]{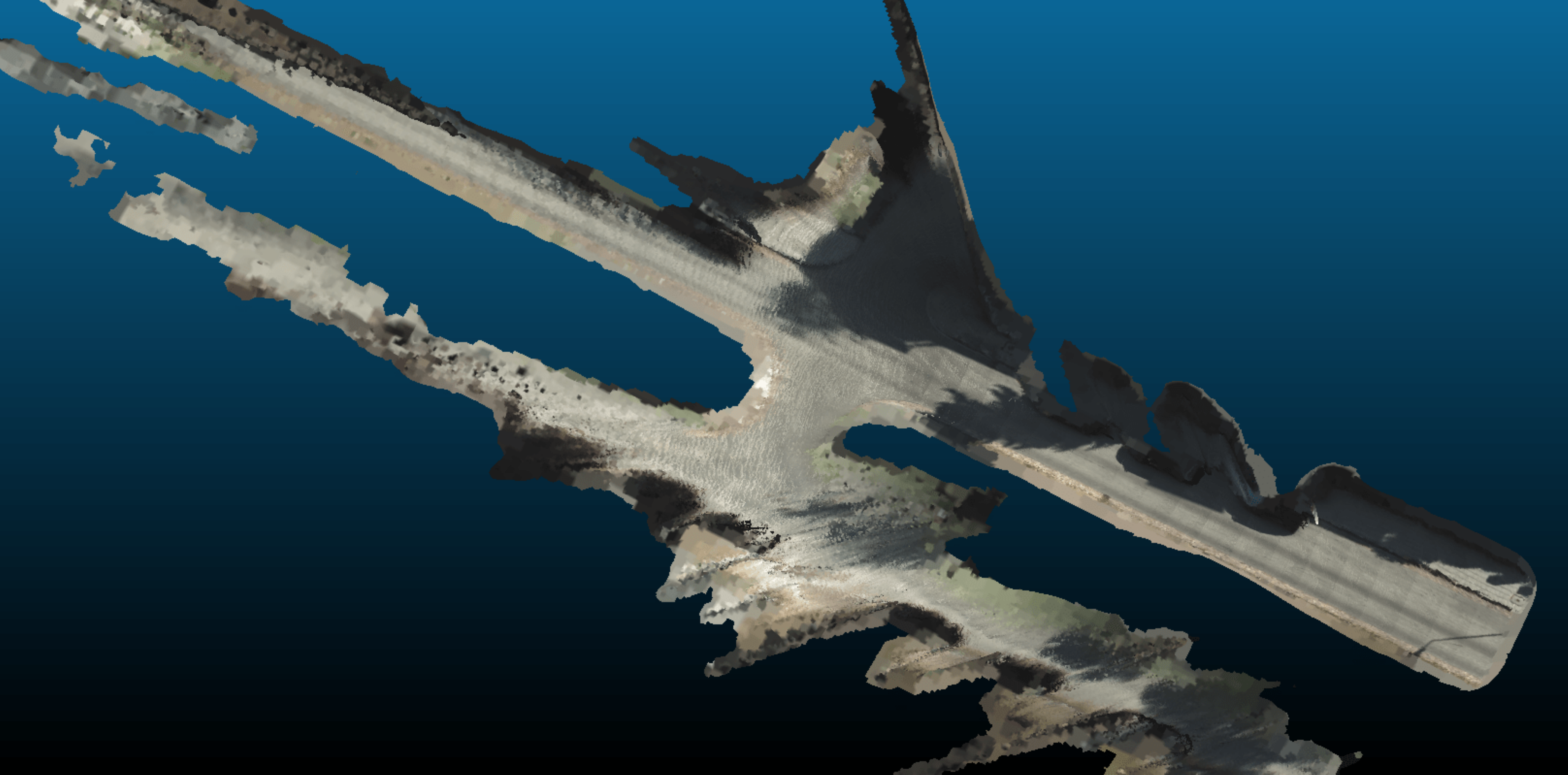} &
		\includegraphics[width=0.65\columnwidth ]{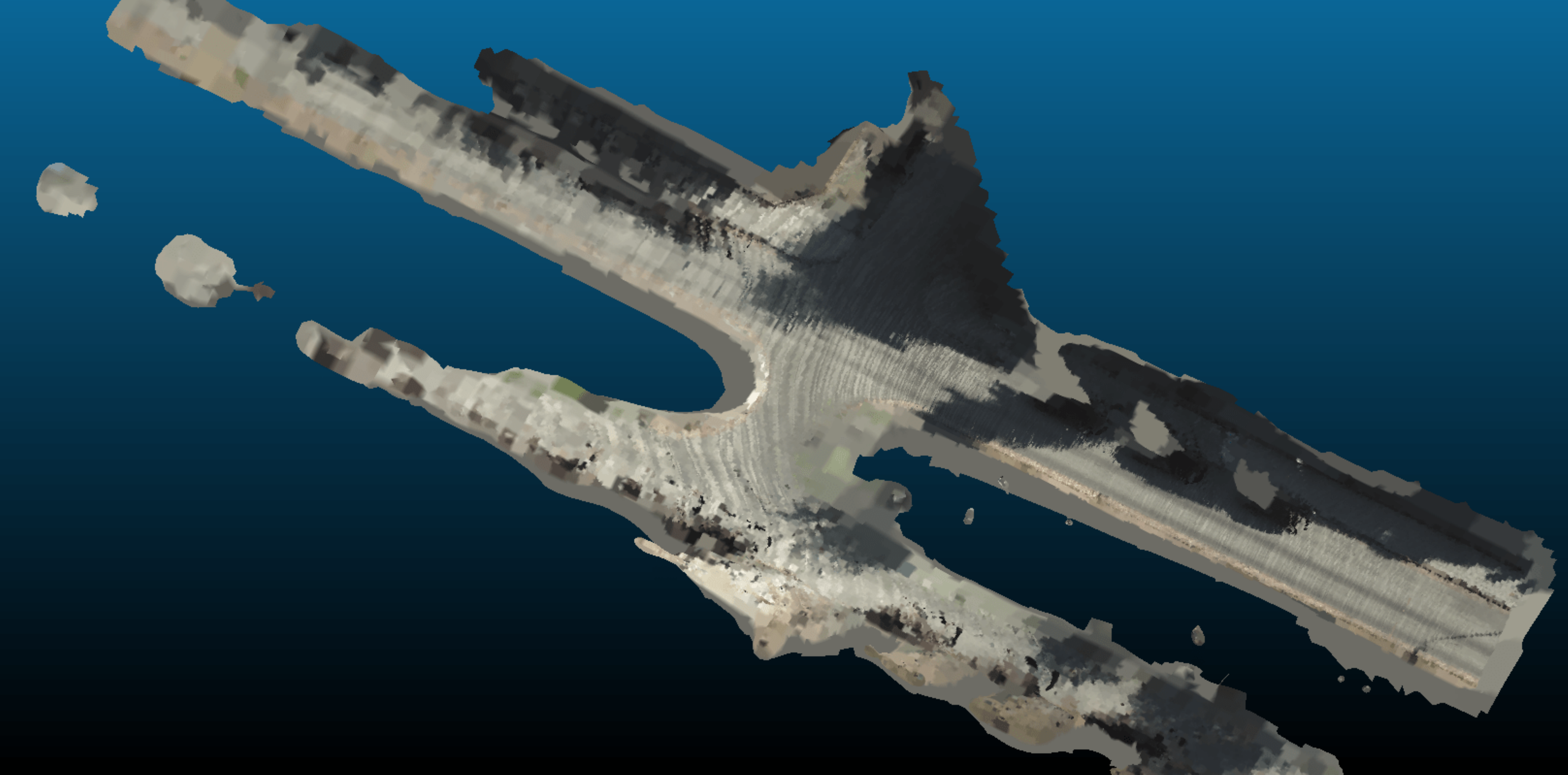} \\

		\includegraphics[width=0.65\columnwidth ]{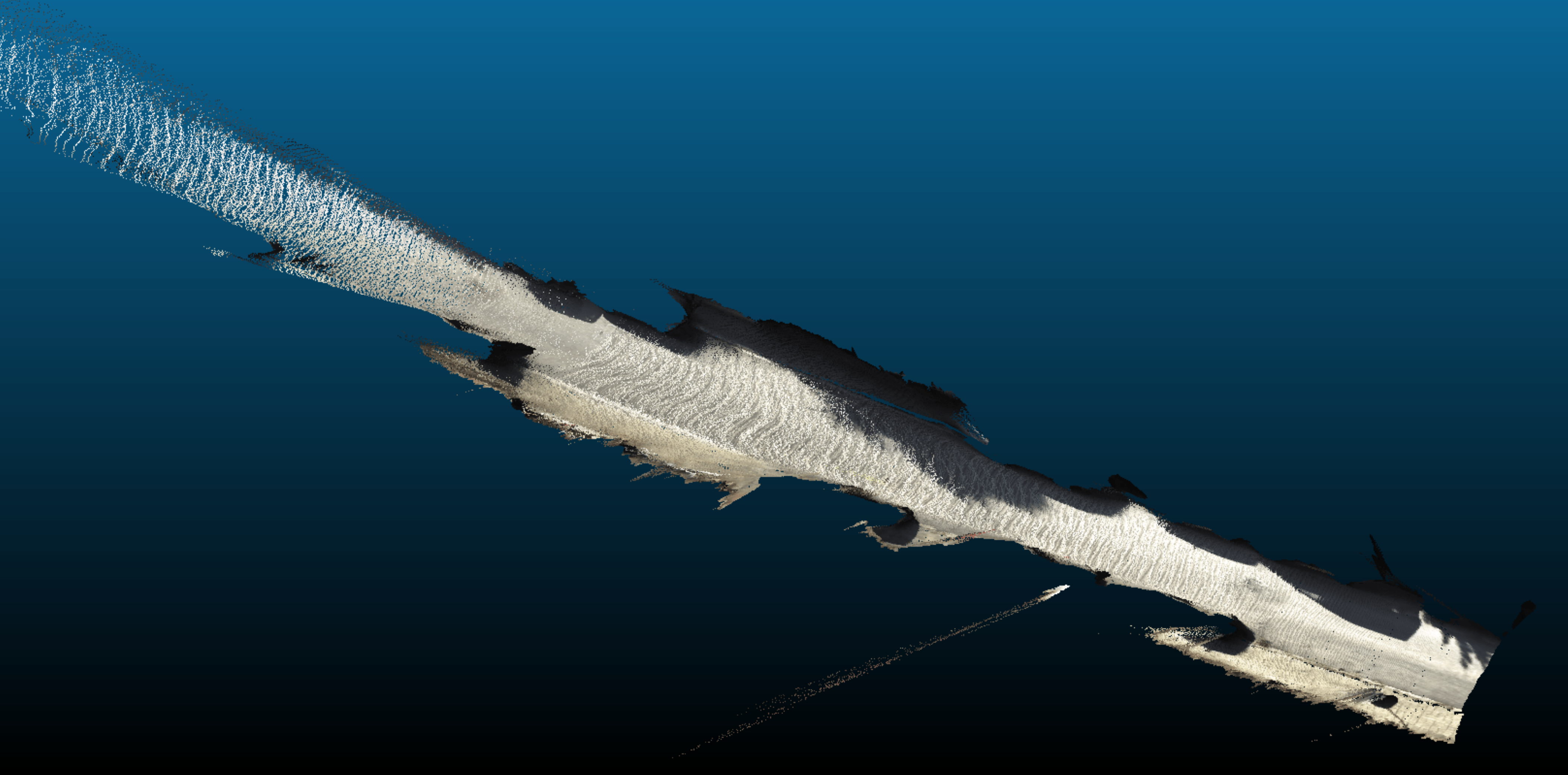} &
		\includegraphics[width=0.65\columnwidth ]{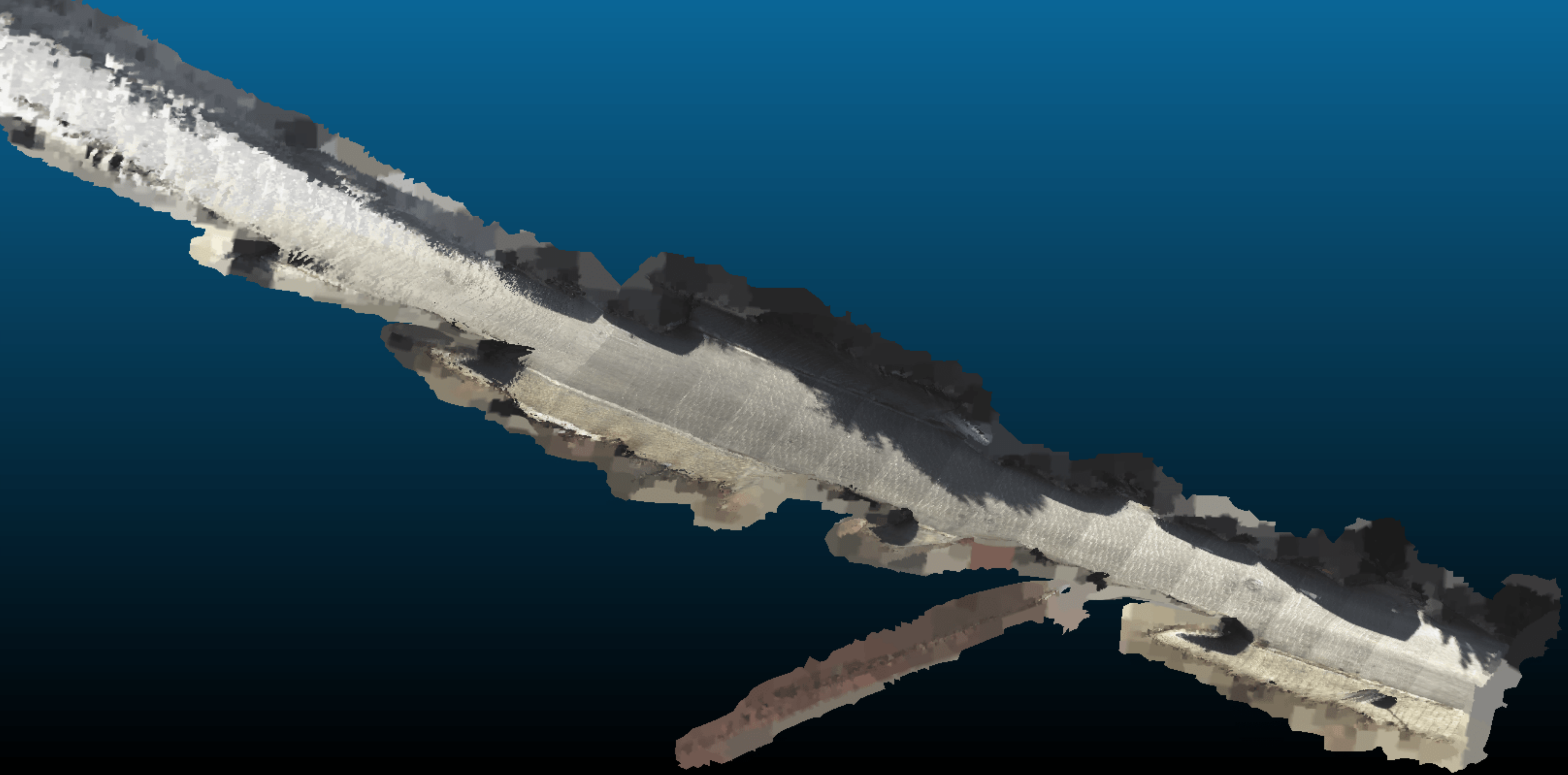} &
		\includegraphics[width=0.65\columnwidth ]{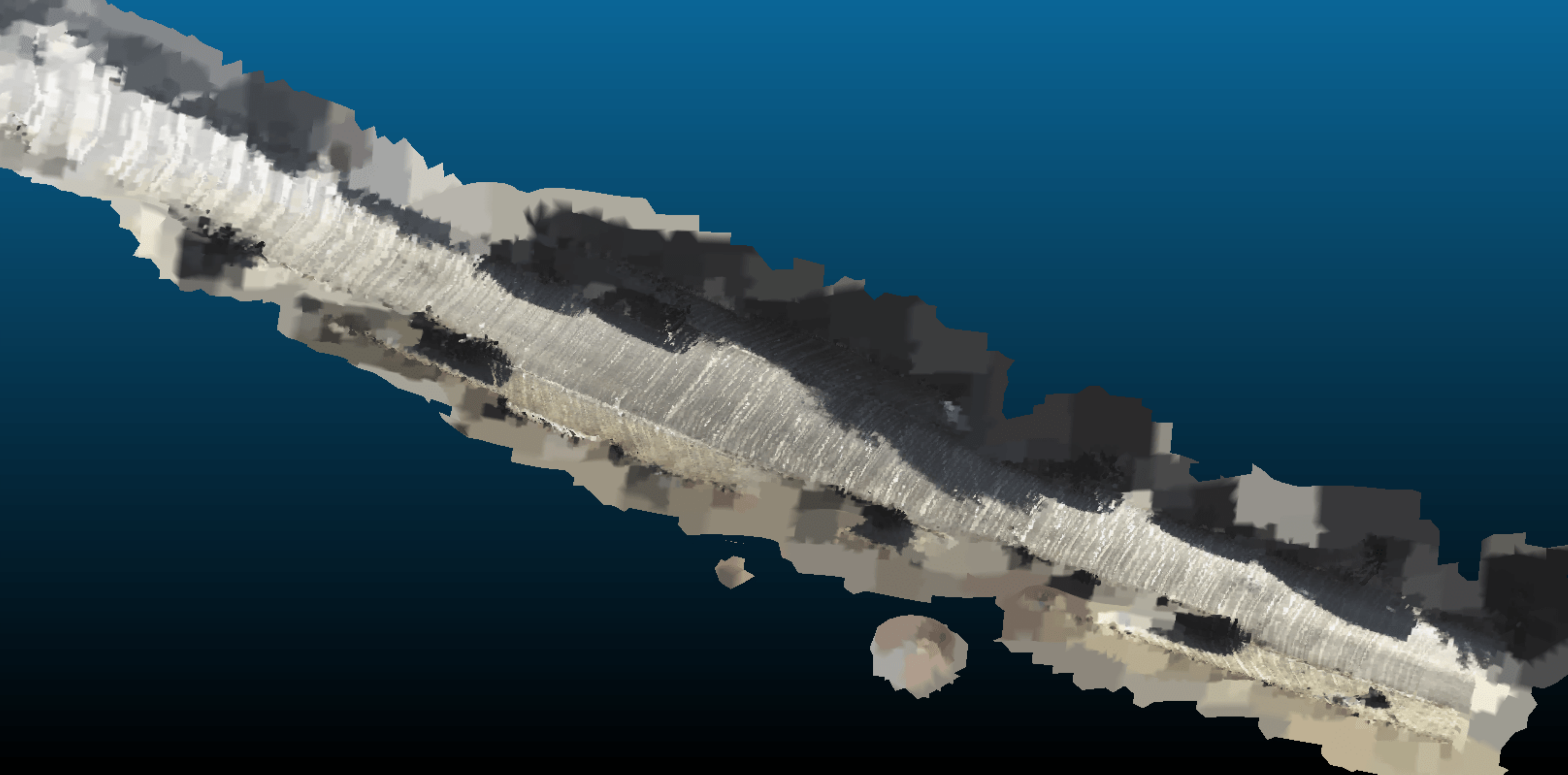} \\

		\includegraphics[width=0.65\columnwidth ]{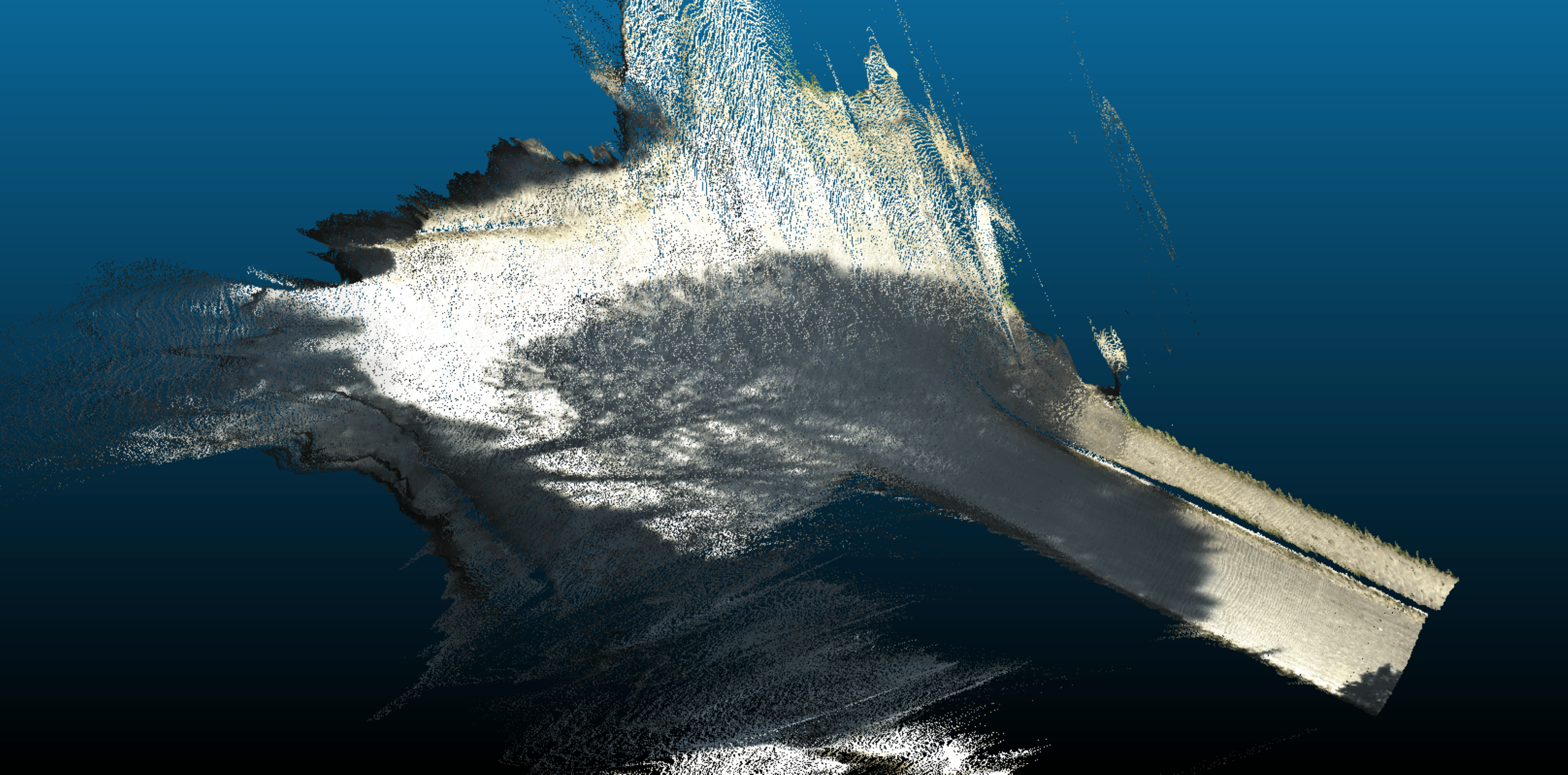} &
		\includegraphics[width=0.65\columnwidth]{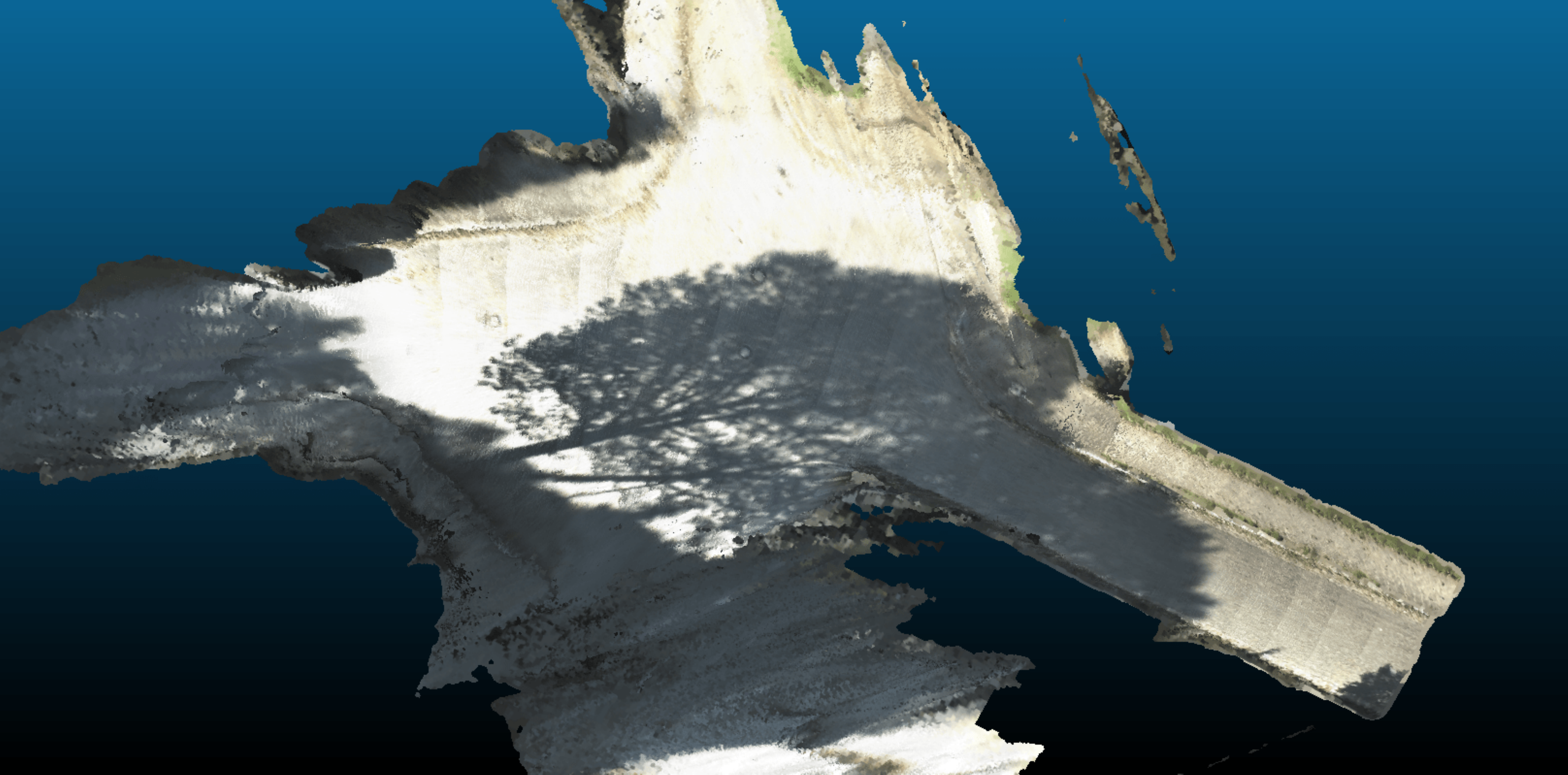} &
		\includegraphics[width=0.65\columnwidth]{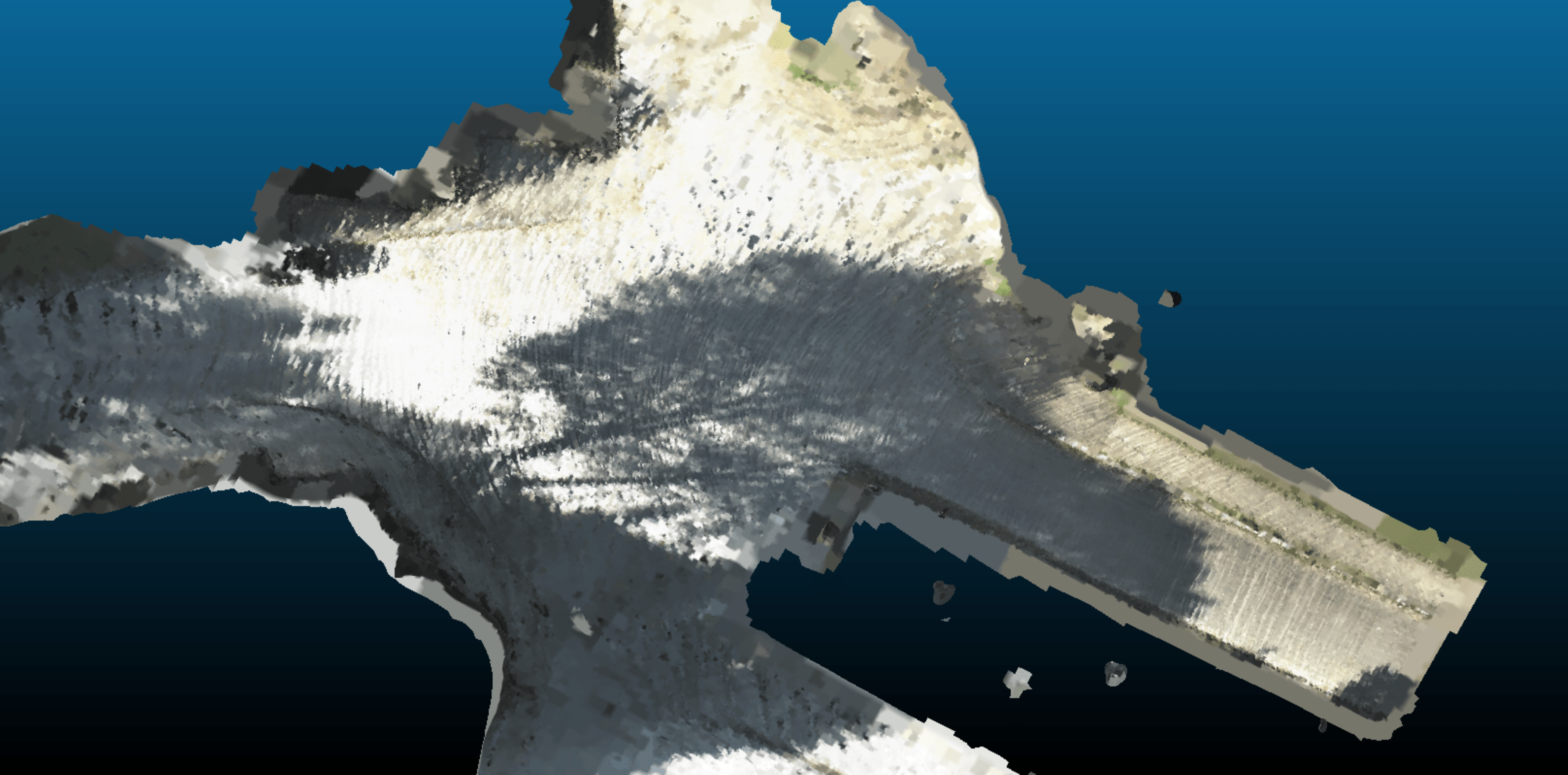} \\
  
	\end{tabular}
 
	\caption{We show geometric quality of our method depicted by predicted point cloud (left) and rendered Poisson mesh (middle) compared to sparse cumulated LiDAR groundtruth (right) on all five scenes of the KITTI-360 dataset.}
	\label{fig:mesh_pcd}
\end{figure*}

\begin{figure*}
	\centering
	\centering
	\setlength{\tabcolsep}{0.008\linewidth}
	\renewcommand{\arraystretch}{0.8}
	\begin{tabular}{c c c c}
            \rotatebox{90}{~~~~~~~~~~None} & \includegraphics[width=0.63\columnwidth]{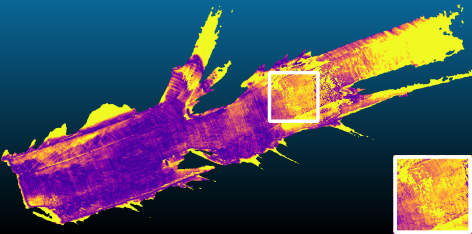} & \includegraphics[width=0.63\columnwidth]{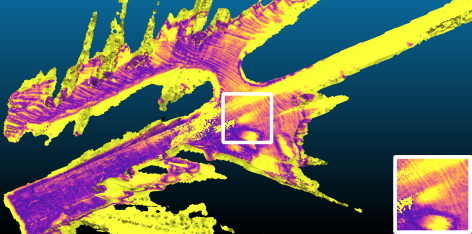} & \includegraphics[width=0.63\columnwidth]{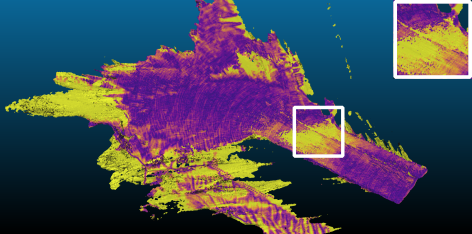} \\
    
            \rotatebox{90}{$\mathcal{L}_{\text{dSSIM}} + \mathcal{L}_{\text{DS}}^*$~\cite{niemeyer2022regnerf}} & \includegraphics[width=0.63\columnwidth]{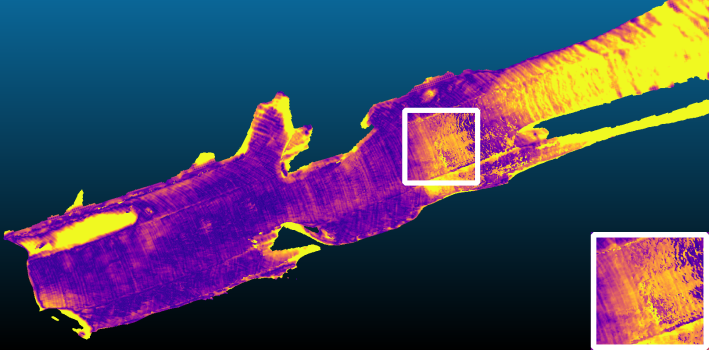} & \includegraphics[width=0.63\columnwidth]{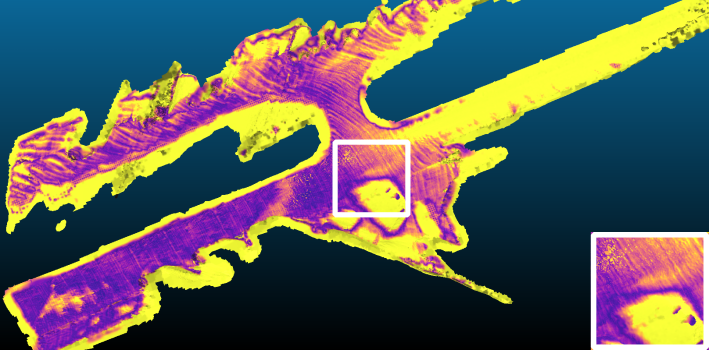} &\includegraphics[width=0.63\columnwidth]{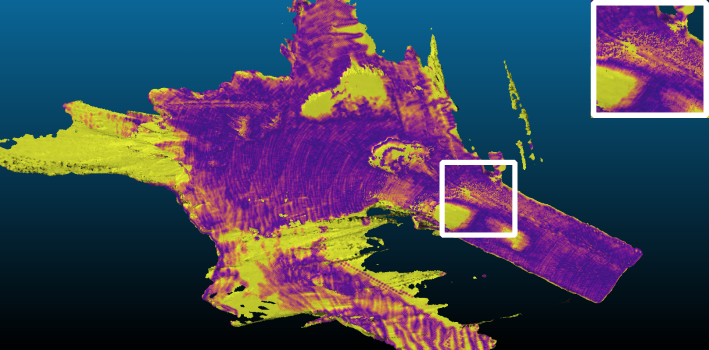} \\
 
            \rotatebox{90}{$\mathcal{L}_{\text{dSSIM}} + \mathcal{L}_{\text{Diff}}^*$~\cite{ehret2022diffnerf}}& \includegraphics[width=0.63\columnwidth]{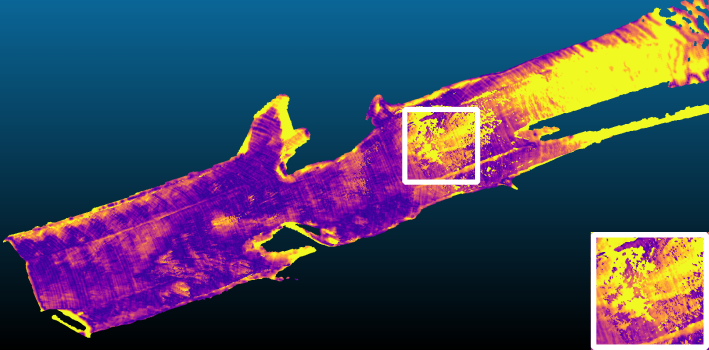} & \includegraphics[width=0.63\columnwidth]{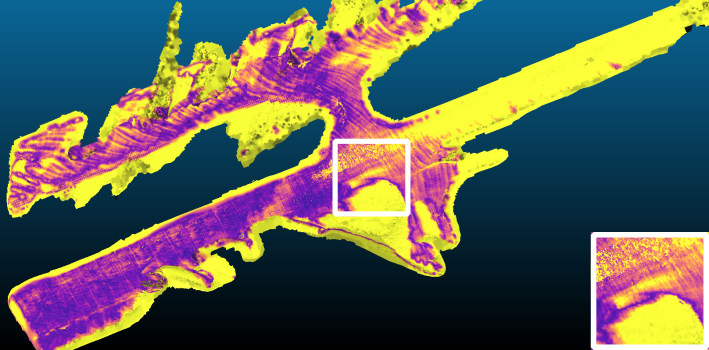} &\includegraphics[width=0.63\columnwidth]{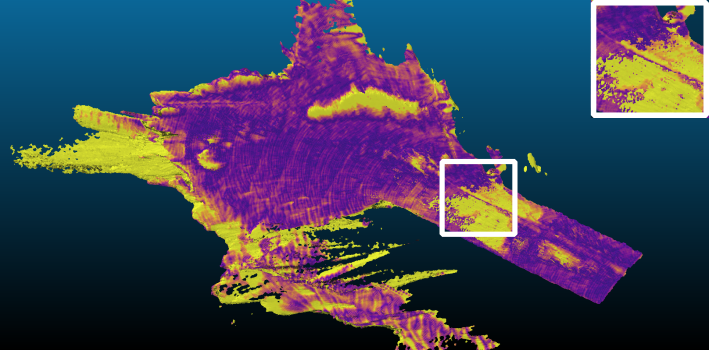} \\
  
            \rotatebox{90}{$\mathcal{L}_{\text{dSSIM}} + \mathcal{L}_{\text{Dist}}^*$~\cite{barron2022mip360}}& \includegraphics[width=0.63\columnwidth]{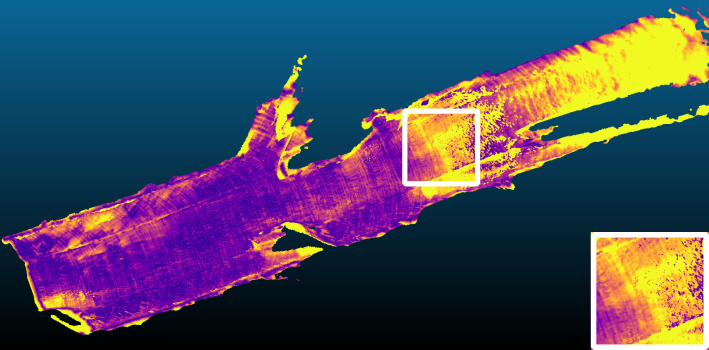} & \includegraphics[width=0.63\columnwidth]{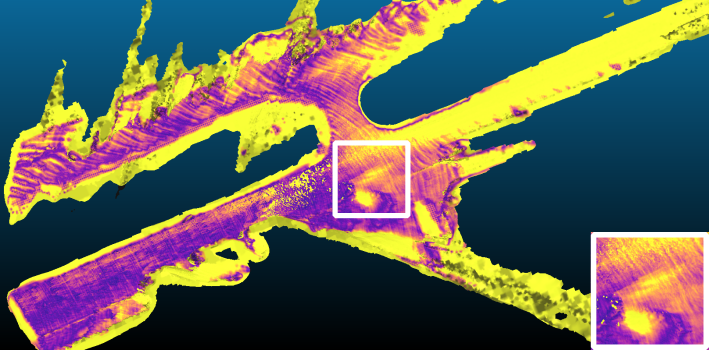} &\includegraphics[width=0.63\columnwidth]{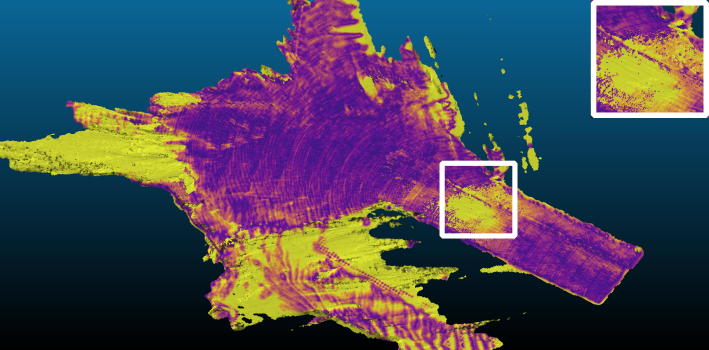} \\
  
            \rotatebox{90}{~~PlaNeRF (ours)}& \includegraphics[width=0.63\columnwidth]{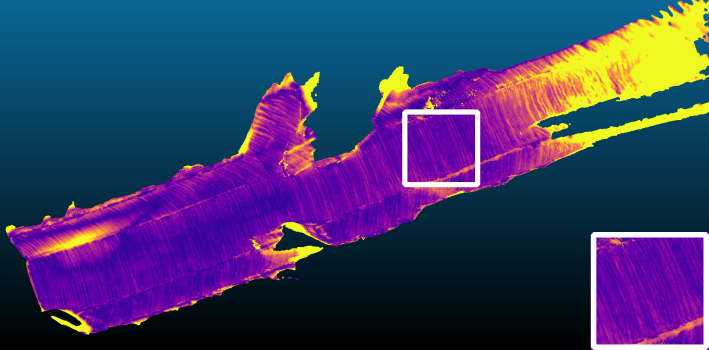} & \includegraphics[width=0.63\columnwidth]{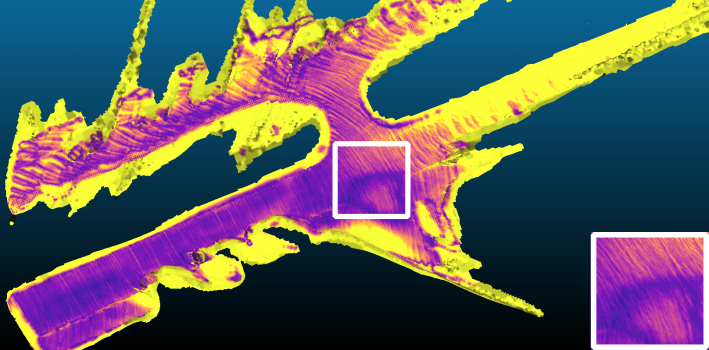} &\includegraphics[width=0.63\columnwidth]{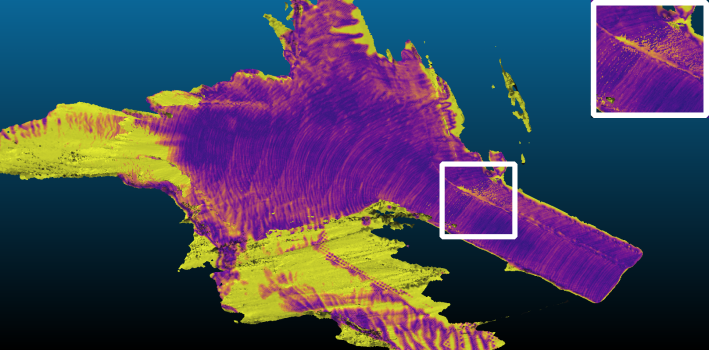} \\
	\end{tabular}
        \caption{Poisson mesh colored by distance to groundtruth mesh by using different regularization strategies on three scenes of the KITTI-360 dataset. As it can be observed, PlaNeRF produces smoother surfaces and more accurate plane geometry when compared to all other regularization methods.}
	\label{fig:3d_error}
\end{figure*}